%% file: main.tex
\documentclass[10pt,twocolumn,letterpaper]{article}
\usepackage[numbers,sort&compress]{natbib}
\PassOptionsToPackage{dvipsnames}{xcolor}
\input{preamble}

\usepackage{enumitem}
\setlist[itemize]{noitemsep,leftmargin=*,topsep=2pt}
\usepackage[hyphens]{url}
\usepackage[hidelinks,unicode]{hyperref}
\hypersetup{
  pdftitle={Rehearse in 3D, Render in 2D: Physically Consistent and Editable Robot World Models},
  pdfauthor={Guangyu Liu, Hanzhang Wang, Huijia Zhang, Hanbei Zhan, Linjie Hu, Bin Xiao, Xuan Zhou, Dong Zhang, Hanwei Liu}
}
\newcommand{\rwInteractionPreservingRehearsal}{3D Interaction Rehearsal}
\newcommand{\rwIPR}{3DIR}
\newcommand{\rwIPRx}{3D Interaction Rehearsal}

\newcommand{\rwInteractionConsistentRehearsal}{Interaction-Consistent Rehearsal}
\newcommand{\rwICR}{ICR}

\newcommand{\rwMotionMaterialHead}{Response-Aware Object Grouping}
\newcommand{\rwMMH}{RAOG}
\newcommand{\rwMMHx}{response-aware object grouping}
\newcommand{\rwMotionMaterialFeature}{visual-motion point features}

\newcommand{\rwGraspMemoryBlock}{Persistent Grasp Memory}
\newcommand{\rwGMB}{PGM}
\newcommand{\rwGMBx}{persistent grasp memory}

\newcommand{\rwContactStructureRefiner}{Contact Structure Refiner}
\newcommand{\rwCSR}{CSR}
\newcommand{\rwCSRx}{contact structure refiner}

\newcommand{\rwRehearsalAlignedVideoGeneration}{Rehearsal-Aligned Video Generation}
\newcommand{\rwRAVG}{RAVG}

\newcommand{\rwRenderAlignedVideoAdaptation}{Render-Aligned Video Adaptation}
\newcommand{\rwRAVA}{RAVA}

\title{Rehearse in 3D, Render in 2D:\\Physically Consistent and Editable Robot World Models}
\author{%
  Guangyu Liu\textsuperscript{1}\quad
  Hanzhang Wang\textsuperscript{2}\quad
  Huijia Zhang\textsuperscript{2}\quad
  Hanbei Zhan\textsuperscript{3}\\
  Linjie Hu\textsuperscript{2}\quad
  Bin Xiao\textsuperscript{2}\quad
  Xuan Zhou\textsuperscript{2}\quad
  Dong Zhang\textsuperscript{2}\quad
  Hanwei Liu\textsuperscript{2,*}}
\date{}
\makeatletter
\renewcommand{\@maketitle}{%
  \begingroup
    \centering
    \vspace*{26pt}
    {\fontsize{16}{19}\selectfont\bfseries \@title\par}
    \vspace{24pt}
    {\fontsize{13}{15}\selectfont \@author\par}
    \vspace{8pt}
    {\fontsize{10.5}{14}\selectfont
      \textsuperscript{1}Tongji University\\
      \textsuperscript{2}Huawei Technologies Ltd.\qquad
      \textsuperscript{3}Zhejiang University\par}
    \vspace{4pt}
    {\small\textsuperscript{*}Corresponding author\par}
  \endgroup
  \vspace{27pt}
  \input{fig/teaser}}
\makeatother
\renewcommand{\dbltopfraction}{0.85}
\renewcommand{\textfraction}{0.10}
\begin{document}
\maketitle
\linespread{0.988}\selectfont
\input{sec/0_abstract}
\input{sec/1_intro}
\input{fig/method_pipeline}
\input{sec/2_related}
\input{sec/4_method}

\input{sec/5_experiments}
\raggedbottom

\input{sec/7_conclusion}

\clearpage
\linespread{1}\selectfont
\par\vspace{6pt}\noindent\begin{minipage}{\columnwidth}
\section*{LLM Usage Statement}
ChatGPT (OpenAI) assisted with manuscript drafting, English translation and language editing, literature organization, and the preparation of figures, tables, and \LaTeX{} files. The authors reviewed and revised the AI-assisted materials and take full responsibility for the final manuscript and its scientific claims.
\end{minipage}\par

{
    \small
    \bibliographystyle{ieeenat_fullname}
    \bibliography{references}
}
\input{supplement}

\end{document}

%% file: preamble.tex
\usepackage{fontspec}
\usepackage{currfile}
\usepackage{caption}
\usepackage{graphicx}
\usepackage{amsmath,amssymb,amsthm}
\usepackage{booktabs}
\usepackage{array}
\usepackage{multirow}
\usepackage{xspace}
\usepackage{tikz}
\usetikzlibrary{arrows.meta,positioning,fit,backgrounds}
\usepackage[dvipsnames]{xcolor}

\definecolor{ResultBest}{RGB}{0,0,0}
\definecolor{ResultSecond}{RGB}{0,0,0}
\newcommand{\bestscore}[1]{\textcolor{ResultBest}{\textbf{#1}}}
\newcommand{\secondscore}[1]{\textcolor{ResultSecond}{\underline{#1}}}
\newcommand{\best}[1]{\bestscore{#1}}
\newcommand{\second}[1]{\secondscore{#1}}

\newcommand{\ours}{RehearsalWorld\xspace}

\newcommand{\x}{\mathbf{x}}

\usepackage{placeins}


%% file: fig/teaser.tex
\par\vspace{6pt}
\begin{minipage}{\textwidth}
\centering
\captionsetup{skip=5pt}
\includegraphics[width=\linewidth]{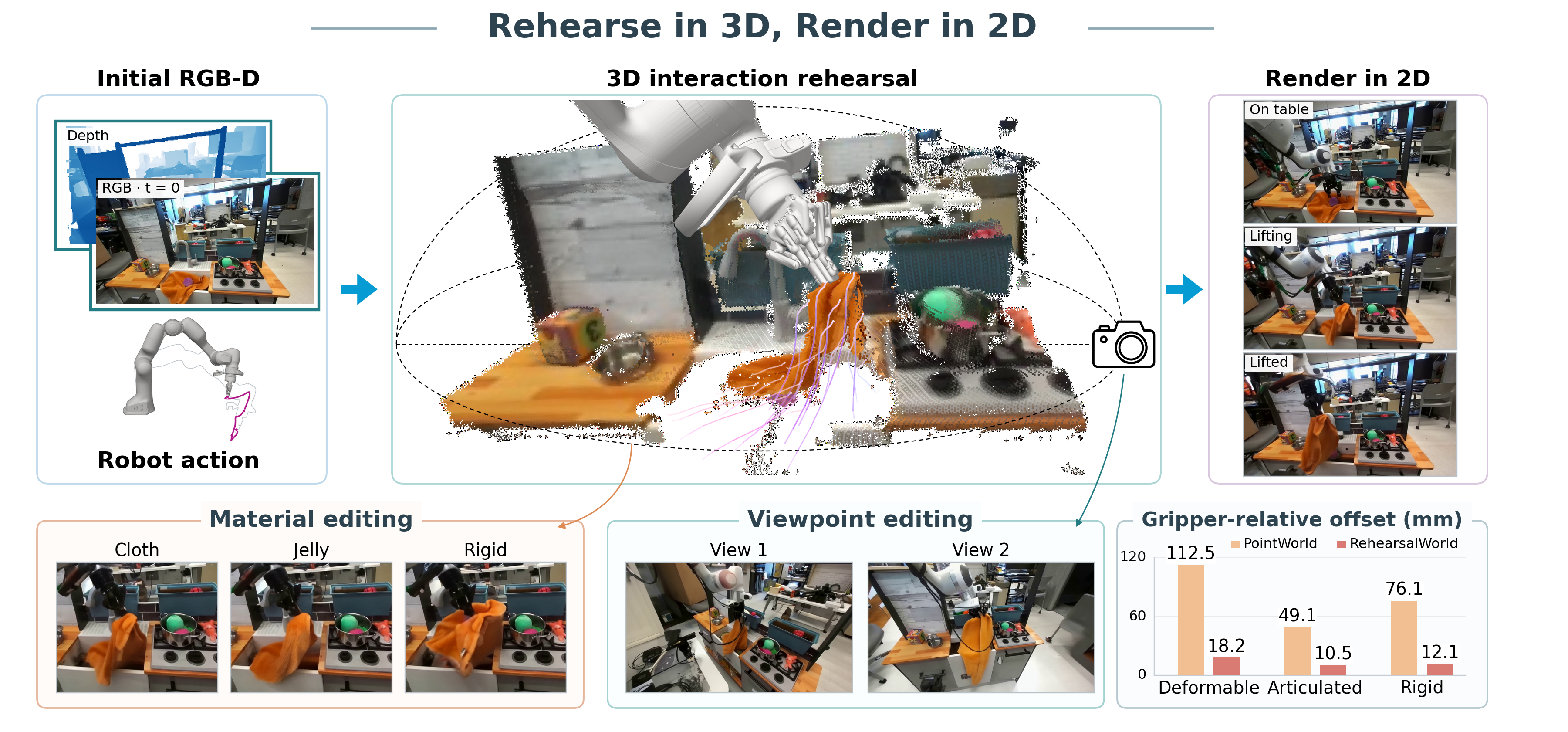}
\captionof{figure}{\textbf{Rehearse in 3D, render in 2D.} From an initial RGB-D observation and robot actions, \ours rehearses interactions in 3D and renders videos, enabling material and viewpoint editing. The bar chart compares gripper-relative offsets for PointWorld and \ours across deformable, articulated, and rigid objects.}
\label{fig:teaser}
\end{minipage}
\par\vspace{6pt}

%% file: sec/0_abstract.tex
\begin{abstract}
{\color{black}
Embodied world models aim to predict future interactions that are both visually realistic and physically consistent. Yet explicit 3D motion prediction alone does not guarantee that object structure and contact relationships are preserved. 
Beyond motion prediction, 3D representations provide an explicit state space for imposing structural and physical constraints.
Building on this insight, we introduce \ours, an editable robot world model driven by \rwIPRx{} \textcolor{black}{(\rwIPR{})}. 
Given an initial RGB-D observation and robot actions, \textcolor{black}{\rwInteractionConsistentRehearsal{}} (\rwICR) uses a pretrained 3D point-flow model as a motion prior. It refines the predicted trajectories through \rwMMHx, \rwGMBx, and \rwCSRx, preserving scene stability, object structure, and grasp attachments. \rwRehearsalAlignedVideoGeneration{} (\rwRAVG) then uses geometric projections of the rehearsal as temporal conditioning. We fine-tune a video generator on projection–video pairs to synthesize realistic interaction videos.
Thanks to the decoupling of physical response from visual synthesis, \ours also supports independent editing of material properties and camera viewpoints. 
Compared with PointWorld on real DROID grasping sequences, \ours reduces gripper-relative offset by approximately 79\%–84\% across deformable, articulated, and rigid objects and lowers reference motion error. For video prediction, it achieves a PSNR of 23.92 dB and an FVD of 149.
\par}
\end{abstract}

%% file: sec/1_intro.tex
\section{Introduction}
\label{sec:intro}

{\color{black}
Embodied video world models generate action-conditioned videos of future interactions to support planning, policy evaluation, and data generation~\cite{guo2025ctrlworld,jang2025dreamgen,agarwal2025cosmos,quevedo2025worldgym,yang2024unisim}.
Recent models produce visually realistic interaction videos with coherent motion.
Visual realism, however, does not guarantee physical consistency: generated interactions can still contain implausible object responses~\cite{bansal2025videophy2,motamed2025physicsiq}.
In robotic manipulation, object motion must also be consistent with the given actions, contact relationships, and the objects' physical properties.

Recent work has introduced explicit 3D representations to better ground interaction prediction in physics.
TesserAct~\cite{zhen2025tesseract}, Aether~\cite{zhu2025aether}, 4DNeX~\cite{chen2025_4dnex}, and Kinema4D~\cite{kinema4d2026} jointly model RGB with depth or point maps to improve spatiotemporal consistency between appearance and geometry.
Yet joint training alone does not ensure RGB--geometry alignment, which depends on cross-modal fusion~\cite{chen2025_4dnex}.
Even well-aligned predictions can violate structural or contact constraints.
Kinema4D, for example, notes that its generated environment responses can still exhibit penetration~\cite{kinema4d2026}.

Other methods learn interaction dynamics directly in 3D, using particles or point clouds to predict action-induced displacement and deformation~\cite{huang2025particleformer,zhang2025pgnd,li2019learning,zhang2024adaptigraph}.
For example, the 3D flow world model PointWorld~\cite{huang2026pointworld} represents RGB-D observations and robot actions in a shared 3D point space and predicts future displacements of the initial scene points.
Its pointwise predictions can capture diverse interaction responses but do not explicitly enforce structural, material, or contact constraints.
On real DROID sequences, we observe background drift, excessive object stretching, and grasp slip in these predictions.
In addition, the predicted flow primarily describes the motion of observed surfaces. Generating complete videos requires additional appearance modeling~\cite{jin2026dynavid}.

Explicit 3D representations not only support geometric modeling and motion prediction but, more importantly, provide a state space for imposing physical constraints on object structure and interactions~\cite{jiang2025phystwin}. Prior work has explored video generation guided by physical simulation~\cite{chen2025physgen3d,li2025wonderplay,psivg2026}. We focus instead on how to use motion priors learned from real interactions to construct more physically consistent 3D states under object-level constraints. Such states could ground both interaction prediction and visual synthesis.

To this end, we introduce \ours, an editable robot world model that uses 3D interaction rehearsal to decouple physical response modeling from video generation.
During \rwInteractionPreservingRehearsal{} \textcolor{black}{(\rwIPR{})}, a pretrained 3D point-flow model~\cite{huang2026pointworld} predicts future interaction trajectories from the initial RGB-D observation and robot actions.
\rwInteractionConsistentRehearsal{} (\rwICR) refines the predicted point displacements using \rwMMHx, \rwGMBx, and \rwCSRx. This reduces background drift, excessive object stretching, and grasp slip.
In \rwRehearsalAlignedVideoGeneration{} (\rwRAVG), we project the corrected 3D states into the target view to obtain temporal geometric conditions.
We fine-tune a video generator on paired dynamic projections and recordings to synthesize realistic videos that follow the rehearsed motion and scene layout.

This decoupling also enables independent editing of material properties and camera viewpoints.
Material edits resimulate the interaction with different material properties while keeping the robot action fixed. Viewpoint edits reproject the same 3D interaction.
Both reuse the video generator without retraining.
Compared with PointWorld~\cite{huang2026pointworld} on real DROID~\cite{khazatsky2024droid} grasping sequences, \ours reduces gripper-relative offset by about 79\%--84\% across deformable, articulated, and rigid objects and lowers reference motion error.
On DROID and BEHAVIOR-1K~\cite{li2022behavior1k}, \ours improves video PSNR by 4.61\,dB and reduces FVD by about 43\% relative to Kinema4D~\cite{kinema4d2026}.

\noindent Our contributions are:
\begin{itemize}
	\item We introduce \ours, an editable robot world model that decouples 3D interaction rehearsal from video synthesis.
	\item We propose \textcolor{black}{\rwICR{}} that refines 3D trajectories via persistent grasp memory and material-conditioned constraints, reducing slip and distortion.
	\item We evaluate 3D rehearsal on DROID and BEHAVIOR-1K and test whether gains in grasp preservation, object structure, and motion accuracy carry over to generated videos.
\end{itemize}
\par}

%% file: fig/method_pipeline.tex
\begin{figure*}[t]
\centering
\captionsetup{skip=3pt}
\includegraphics[width=\textwidth]{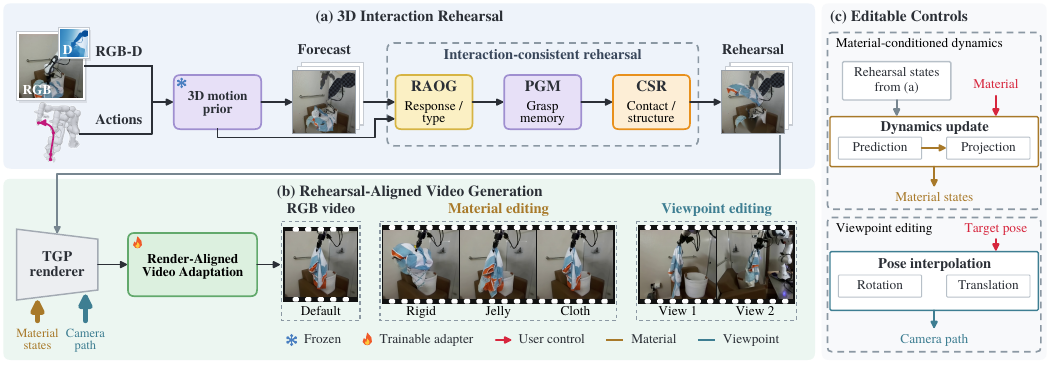}
\caption{\textbf{Overview of \ours.} (a) \rwInteractionPreservingRehearsal{} (\rwIPR{}). (b) \textcolor{black}{\rwRehearsalAlignedVideoGeneration{}} (\rwRAVG{}). (c) Material and viewpoint editing, which share the rendering and generation pipeline in (b).}
\label{fig:method}
\end{figure*}

%% file: sec/2_related.tex
\section{Related Work}
\label{sec:related}

\noindent \textbf{Embodied video world models.} \enspace
UniSim~\cite{yang2024unisim} and Cosmos~\cite{agarwal2025cosmos} learn interaction dynamics from large-scale video data. Ctrl-World~\cite{guo2025ctrlworld} and WorldGym~\cite{quevedo2025worldgym} use generated worlds to evaluate policies, while DreamGen~\cite{jang2025dreamgen} synthesizes robot training data. Physical properties are typically implicit in pixel generation~\cite{bansal2025videophy2,motamed2025physicsiq}, making material and contact constraints difficult to enforce. \ours supplies the video generator with explicitly constrained object states.

\noindent \textbf{3D and 4D world models.} \enspace
Learned particle and graph dynamics~\cite{huang2025particleformer,zhang2024adaptigraph,zhang2025pgnd} predict object motion, while 3D flow foundation models~\cite{huang2026pointworld,ye2026pointcast} learn from large-scale data to predict action-conditioned pointwise motion. PointZero~\cite{duisterhof2026pointzero} pretrains transferable dynamics through sparse-to-dense 3D trajectory completion. TesserAct~\cite{zhen2025tesseract}, Aether~\cite{zhu2025aether}, 4DNeX~\cite{chen2025_4dnex}, and Kinema4D~\cite{kinema4d2026} jointly generate RGB and geometry. An explicit geometric representation alone does not impose material or grasp constraints. To determine how frozen motion proposals should be refined, \ours augments them with persistent object identities, response types, and attachment states.

\noindent \textbf{Physical constraints and 3D-guided video generation.} \enspace
Physics-based digital twins estimate parameters from video and simulate interactions~\cite{jiang2025phystwin,li2023pacnerf,xie2024physgaussian}. PhysGen3D~\cite{chen2025physgen3d}, WonderPlay~\cite{li2025wonderplay}, and PSIVG~\cite{psivg2026} combine simulation with video generation; PhysCtrl~\cite{wang2025physctrl} learns controllable physical motion. Diffusion as Shader~\cite{gu2025das} and GEN3C~\cite{ren2025gen3c} condition synthesis on 3D motion or scene caches. \ours applies object constraints to refine action-conditioned trajectories learned from real interactions. The refined trajectories guide material editing and render-aligned video synthesis.

%% file: sec/4_method.tex
\section{Method}
\label{sec:method}
\label{sec:overview}
\raggedbottom

Given an initial RGB-D observation and robot actions, \ours generates visually realistic and interaction-consistent future videos as shown in Fig.~\ref{fig:method}. During \rwInteractionPreservingRehearsal{} \textcolor{black}{(\rwIPR{})}, \ours uses a pretrained 3D point-flow model as a motion prior and applies the \textcolor{black}{\rwInteractionConsistentRehearsal{}} (\rwICR) \textcolor{black}{component} to refine scene point trajectories under grasp and structural constraints. In \rwRehearsalAlignedVideoGeneration{} \textcolor{black}{(\rwRAVG{})}, it projects the refined trajectories along a specified camera path and generates videos from the target viewpoints. The pipeline also supports material and viewpoint editing, enabling control over object dynamics and camera perspective.

\subsection{Task and State Representation}
\label{sec:formulation}
Given an initial RGB-D observation $(I_0,D_0)$, observation camera calibration $C_{\mathrm{obs}}$, and robot actions $q_{0:T}$, \ours predicts a refined 3D state sequence $\hat Z_{0:T}$ and generates an RGB video $I_{0:T}^{C}$ along a specified camera path, where $T$ is the prediction horizon.

We represent the scene state at time $t$ as $Z_t=\{z_t^j\}_{j=1}^{N}$, where $z_t^j$ is the candidate position of point $j$ in world coordinates. We denote the 3D interaction trajectories by 3D state sequence $Z_{0:T}=\{Z_t\}_{t=0}^{T}$ and the refined trajectories by $\hat Z_{0:T}$. All point positions are expressed in world coordinates and retain the same indices throughout prediction.

We model robot geometry separately using URDF forward kinematics.
To describe gripper--object interactions, we define a local gripper coordinate frame and denote its transformation to the world frame at time $t$ by $T_t^g$.
The camera path is $C=\{C_t\}_{t=0}^{T}$, which contains the fixed camera intrinsics and world-to-camera extrinsic transformation. The path starts at the observation viewpoint, $C_0=C_{\mathrm{obs}}$, and the camera remains fixed by default.


\subsection{\rwInteractionPreservingRehearsal}
\label{sec:rehearse}
\label{sec:separation}
\label{sec:state}
\label{sec:latent}
\label{sec:diagnosis}
\label{sec:decoder}
\label{sec:projection}

We firstly utilize a pretrained 3D point-flow model $F$~\cite{huang2026pointworld} to predict the future 3D interaction trajectories of scene points from the initial RGB-D $(I_0,D_0)$ and robot actions $q_{0:T}$. To suppress background drift, excessive object stretching, and grasp slippage in pointwise predictions, we introduce \textcolor{black}{\rwICR{}}, which refines scene point displacements using object membership, grasp references, and local structural constraints,
\begin{equation}
\begin{aligned}
Z_{0:T},\mathbf f&=F(I_0,D_0;q_{0:T}),\\
\hat Z_{0:T}&=\mathcal U(Z_{0:T}, \mathbf f;q_{0:T}).
\end{aligned}
\label{eq:interaction_rehearsal}
\end{equation}
where $\mathcal U$ is the \rwICR{} refinement operator, and $\mathbf f$ denotes the \rwMotionMaterialFeature{} extracted from the point-flow model.
\rwICR{} applies object-level constraints to scene point displacements through three connected components (Fig.~\ref{fig:method}(a)).
The \rwMotionMaterialHead{} (\rwMMH) identifies point groups for responsive objects and anchors the static background.
The \rwGraspMemoryBlock{} (\rwGMB) establishes attachments between responsive objects and the gripper.
The \rwContactStructureRefiner{} (\rwCSR) uses object groups, object types, and attachment information to apply contact and structural constraints to the predicted point displacements, reducing grasp slip and structural distortion.

\begin{figure}[!t]
\centering
\captionsetup{skip=3pt}
\includegraphics[width=\columnwidth]{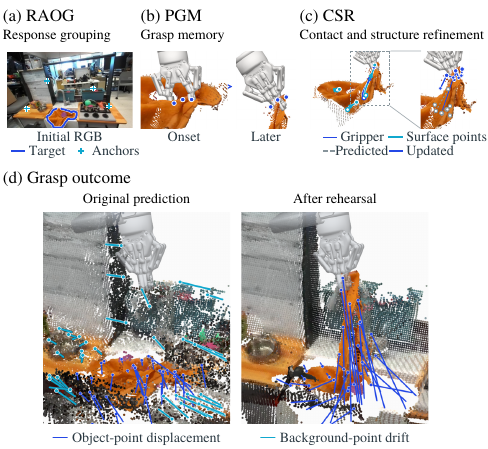}
\caption{\textbf{ICR components and their effect on grasping.} (a) \rwMMH{} groups points on responsive objects. (b) \rwGMB{} stores and reuses grasp references. (c) \rwCSR{} at \textcolor{black}{$t=60$ (4.07\,s)} shows the original prediction (dashed gray), refined trajectories and attachment points (blue). (d) The gripper lifts the towel after rehearsal.}
\label{fig:loop}
\end{figure}

\noindent\textbf{\rwMotionMaterialHead{} (\rwMMH{}).}\enspace
\textcolor{black}{\rwMMH{}} estimates how each scene point responds to the given robot actions, partitions the scene into responsive object point groups and static background points, and predicts each responsive object's physical type. The object groups support subsequent grasp and structural refinement, while background points remain at their initial observed positions to suppress drift.

We extract intermediate \rwMotionMaterialFeature{} $\mathbf f\in\mathbb R^{N\times256}$ from the point-flow model and apply mean and max pooling across points. For each point $j$, we apply an exponential transformation to the pooled features together with the positional encoding of its initial coordinates, then flatten the result and concatenate it with $\mathbf f^j$ to obtain the augmented feature $\hat{\mathbf f}^j\in\mathbb R^{768}$. Then, we utilize an MLP to predict response scores, which are then thresholded to separate responsive points $\mathcal P$ from background points $B$.

We cluster the responsive points $\mathcal P$ into $M$ disjoint object point groups $\mathcal P^i$ using distances between initial point positions and differences in predicted displacement~\cite{song2024geometryconsistency}. The \rwMotionMaterialFeature{} within each group are used to predict its type $\kappa^i$, such as rigid, deformable, or articulated, which determines the structural and material constraints. This gives the object set $\mathcal O=\{(\mathcal P^i,\kappa^i)\}_{i=1}^{M}$. Object membership remains fixed throughout prediction, and background points are held at their initial positions: $\widehat z_t^j=z_0^j$, $j\in B$. The detailed architecture of \rwMMH{} is provided in the supplementary material.



\noindent\textbf{\rwGraspMemoryBlock{} (\rwGMB{}).}\enspace
To reduce object slip after a grasp is established, \rwGMB{} stores object point positions relative to the gripper when it detects grasp onset.
\rwCSR{} then uses these stored positions and the current gripper pose to determine target positions for contact points~\cite{longhini2025clothsplatting}.

\rwGMB{} identifies the grasp onset time $t_c^i$ for object $O^i$ from gripper--object proximity and the gripper's open/closed state, and tracks attachment with $b_t^i\in\{0,1\}$. At grasp onset, it selects a set of contact points $\mathcal A^i$ from $\mathcal P^i$ and stores the positions of all points in the object group in the gripper frame:
\begin{equation}
r_j^i=(T_{t_c^i}^{g})^{-1}z_{t_c^i}^{j},
\qquad j\in\mathcal P^i.
\label{eq:pgm_grasp_reference}
\end{equation}
At the same time, \rwGMB{} constructs a neighborhood graph $\mathcal E^i$ within the object and records reference edge lengths $\ell_{jk}^{c,i}$.
The attachment state $b_t^i$, contact point set $\mathcal A^i$, and local coordinates $\{r_j^i\}_{j\in\mathcal P^i}$ form the interaction memory $\mathcal H_t^i$, which \rwCSR{} uses to set contact targets and apply structural constraints.

\noindent \textbf{\rwContactStructureRefiner{} (\rwCSR).} \enspace
We introduce the \textcolor{black}{\rwCSR{}} to improve the physical consistency of 3D interaction rehearsal. Given the object type $\kappa^i$, it selects a configuration $\mathcal Q_{\kappa^i}$ and refines candidate trajectories using the grasp information stored by \rwGMB{}, subject to structural and contact constraints. We organize these material-conditioned constraints into three groups: internal structure, environmental interactions, and physical effects.

\noindent\hangindent=1em\hangafter=1\makebox[1em][l]{\textbullet}\textcolor{black}{\textbf{Internal structural constraints}} preserve geometric relationships within each object using the local structure recorded by \rwGMB{} as a reference. Rigid fitting preserves the shape of rigid objects. For deformable objects, neighboring-edge constraints $\|z_t^j-z_t^k\|\leq(1+\epsilon_{\kappa^i})\ell_{jk}^{c,i}$ limit excessive stretching, where $\epsilon_{\kappa^i}$ is the allowed relative extension~\cite{muller2007pbd}.

\noindent\hangindent=1em\hangafter=1\makebox[1em][l]{\textbullet}\textcolor{black}{\textbf{Environmental interaction constraints}} \textcolor{black}{keep objects attached to the gripper.} During a grasp, we set $z_{t,\mathrm{target}}^j=T_t^g r_j^i$ for $j\in\mathcal A^i$ so that contact points follow the gripper. We adjust object points based on their distance to the sampled robot surface to limit penetration~\cite{macklin2014unified}.


\noindent\hangindent=1em\hangafter=1\makebox[1em][l]{\textbullet}\textcolor{black}{\textbf{Physical-effect constraints}} approximate the effects of gravity, environmental support, and contact friction on object motion. For free points that meet the lifting condition, we set $h_j^+=\max\{h_{\min,j}^i,h_j-\Delta h_j\}$, where $\Delta h_j$ decays over iterations and the support bound $h_{\min,j}^i$ is derived from the height at grasp onset.


Let $Z_t^i=(z_t^j)_{j\in\mathcal P^i}$ denote the candidate object coordinates in point-index order. Using interaction memory, robot state, and the static scene, \rwCSR{} accumulates corrections from the three constraint groups according to the selected configuration to obtain
\begin{equation}
\widehat Z_t^i=Z_t^i
+\Delta Z_{t,\mathrm{in}}^i
+\Delta Z_{t,\mathrm{out}}^i
+\Delta Z_{t,\mathrm{phys}}^i.
\label{eq:csc_state_update}
\end{equation}
The supplementary material provides details on the configurations, parameters, and solver, along with additional ablation analyses.

We preserve $\widehat Z_0=Z_0$ and merge the refined object points with the background points anchored by \rwMMH{}, retaining their original indices. The resulting 3D interaction trajectories $\widehat Z_{0:T}$ condition video generation on interaction motion and scene layout. Fig.~\ref{fig:loop}(a)--(c) shows contact preservation and structural refinement during grasping and lifting, while Fig.~\ref{fig:loop}(d) shows reduced background drift, excessive stretching, and grasp slip after refinement, improving the physical consistency of 3D interaction rehearsal.



\subsection{Rehearsal-Aligned Video Generation}
\label{sec:render}
To generate visually realistic videos of future interactions, temporal geometry projection (TGP) converts the refined 3D rehearsal trajectories into a sequence of coarse renders from the target viewpoints. \textcolor{black}{\rwRenderAlignedVideoAdaptation{}} (\textcolor{black}{\rwRAVA{}}) then generates interaction videos that follow the rehearsed motion.

\noindent\textbf{Temporal geometry projection (TGP).}\enspace
TGP transfers point displacements to object surfaces via locally weighted skinning. Let $S_0$ denote the scene surface reconstructed from the initial RGB-D observation, including its RGB appearance and skinning bindings. We render the updated object surfaces, static scene, and robot geometry $G_t$ along camera path $C$ to obtain
\begin{equation}
R_t^C=\operatorname{Render}_{C_t}(\widehat Z_t,G_t;S_0),
\qquad t=1,\ldots,T.
\label{eq:temporal_geometry_projection}
\end{equation}
The coarse RGB sequence $R_{1:T}^C$ provides motion, layout, and occlusion cues in the target views.


\noindent \textbf{\textcolor{black}{\rwRenderAlignedVideoAdaptation{}} (\textcolor{black}{\rwRAVA{}}).} \enspace
We extend FixAnything's render conditioning~\cite{vuong2026fixanything} to dynamic projections of robot--object interactions.
A VAE encodes the coarse renders as temporal conditioning, and the observed first frame $I_0$ serves as the sole trusted appearance reference.
The generator $H_\phi$ produces the RGB video as
\begin{equation}
\begin{aligned}
R_0^C = I_0, \quad I_{0:T}^C = H_\phi(I_0,R_{0:T}^C).
\end{aligned}
\label{eq:render_generate}
\end{equation}

During \textcolor{black}{the training phase}, we drive the same TGP module with reference 3D trajectories and train the generator on pairs of coarse renders and real or simulated RGB frames matched in time and viewpoint. Except for the first frame, paired RGB frames serve only as supervision and are not input conditions. During \textcolor{black}{the inference phase}, we drive TGP with the 3D rehearsal trajectories to construct the render conditions, using the initial RGB-D frame as the input.


\subsection{Material and Viewpoint Editing}
\label{sec:edit}
By decoupling 3D motion from visual appearance, \ours can vary an object's response to the same action or show the same interaction from new viewpoints. These two forms of editing modify the 3D state input to TGP and its camera path, respectively (Fig.~\ref{fig:method}(c)).

\noindent \textbf{Material-conditioned dynamics (MCD).} \enspace
MCD compares material responses under the same initial state and robot actions.
We sample dense particles from the observed surface of $O^i$ and track their positions $Y^i$ and velocities $V^i$.
The editable parameters $\theta^i$ control shape preservation, stretch resistance, damping, and long-range attachment. The class $\kappa^i$ selects the default rehearsal constraints.

At each substep, velocities and gravity give candidate positions $\widetilde Y^i$, which we update using position-based dynamics~\cite{muller2007pbd} as follows:
\begin{equation}
Y^{i,+}=\mathcal P_{\theta^i,\mathcal A^{\mathrm{mat},i},G_t}
\bigl(\widetilde Y^i;\bar X^i\bigr).
\label{eq:material_edit_update}
\end{equation}
Here, $\mathcal A^{\mathrm{mat},i}$ specifies action-driven contacts, and $\bar X^i$ provides weak motion guidance from the refined rehearsal.
We pass the resulting object trajectories to TGP along with the rest of the scene and robot geometry. Fig.~\ref{fig:material_tracks} compares three corresponding point trajectories and generated RGB frames across materials.

\noindent \textbf{Viewpoint editing.} \enspace
\label{sec:viewpoint}
Viewpoint editing changes only the camera path $C$, keeping the 3D trajectories and contact sequence fixed. The path starts at the observation viewpoint and gradually transitions to the requested view. TGP recomputes projections and occlusions at each time step, and the video generator produces the new-view video using Eq.~\eqref{eq:render_generate}, with the real first frame still serving as the only trusted frame. 

Both edits can be combined by projecting the material-edited states along a specified camera path. They share the same TGP and video generator, with no retraining for individual edits.

\input{fig/experimental_floats}

%% file: fig/experimental_floats.tex
\begin{figure}[!tbp]
\centering
\captionsetup{skip=3pt}
\includegraphics[width=\columnwidth]{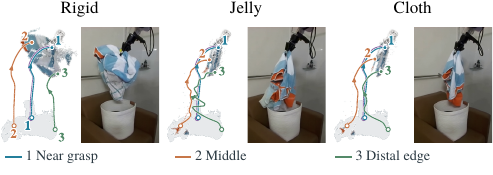}
\caption{\textbf{Material-dependent motion.} With fixed observations and actions, three corresponding points trace different 3D paths. Left: trajectories at a shared viewpoint and scale. Right: generated RGB crops at \textcolor{black}{$t=60$ (4.07\,s)}.}
\label{fig:material_tracks}
\end{figure}

\input{fig/category_results_table}

\input{fig/route_results_table}

\input{fig/qualitative_comparison}

\begin{figure}[!t]
\centering
\includegraphics[width=\columnwidth,trim=0 12bp 0 0,clip]{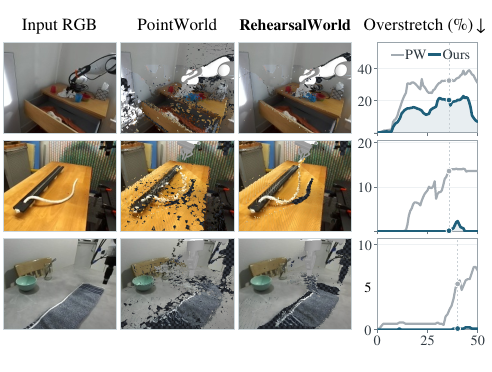}
\caption{\textbf{Structural preservation in 3D rehearsal.}
Point clouds from PointWorld and \ours{} are compared on the left.
The right plots show the fraction of overstretched point pairs, with dashed
lines marking the displayed frames.}
\label{fig:phys}
\vspace{4.935pt}
\end{figure}

%% file: fig/category_results_table.tex
\begin{table*}[t]
	\caption{\textbf{Native 3D interaction rehearsal on DROID.}
		D, A, and R denote deformable, articulated, and rigid objects.
		Offsets and position errors are in mm, and overstretching rates are in \%.
		Gripper-relative offset, the three position errors, and the overstretching rate
		assess grasp stability, trajectory accuracy, and structural preservation,
		respectively.}
	\label{tab:categories}
	\centering
	\color{black}
	\fontsize{8}{9.5}\selectfont
	\setlength{\tabcolsep}{1.3pt}
	\renewcommand{\arraystretch}{1.08}
	
	\newcommand{\rwDroidMethodWidth}{0.17\textwidth}
	\newcommand{\rwDroidValueWidth}{%
		\dimexpr(\textwidth-\rwDroidMethodWidth-30\tabcolsep)/15\relax}
	\newcommand{\rwDroidMetricWidth}{%
		\dimexpr\rwDroidValueWidth*3+4\tabcolsep\relax}
	\newcommand{\rwDroidMetricHeader}[1]{%
		\parbox[c]{\rwDroidMetricWidth}{\centering #1}}
	
	\begin{tabular}{@{}>{\raggedright\arraybackslash}p{\rwDroidMethodWidth}
			*{15}{>{\centering\arraybackslash}p{\rwDroidValueWidth}}@{}}
		\toprule
		\multirow{2}{*}{Method}
		& \multicolumn{3}{c}{\rwDroidMetricHeader{Gripper-relative offset$\downarrow$}}
		& \multicolumn{3}{c}{\rwDroidMetricHeader{Pre-grasp error$\downarrow$}}
		& \multicolumn{3}{c}{\rwDroidMetricHeader{Moving-point error$\downarrow$}}
		& \multicolumn{3}{c}{\rwDroidMetricHeader{Post-grasp\\free-point error$\downarrow$}}
		& \multicolumn{3}{c@{}}{\rwDroidMetricHeader{Overstretching rate$\downarrow$}} \\
		\cmidrule(lr){2-4}\cmidrule(lr){5-7}\cmidrule(lr){8-10}
		\cmidrule(lr){11-13}\cmidrule(l){14-16}
		& D & A & R & D & A & R & D & A & R & D & A & R & D & A & R \\
		\midrule
		PointWorld~\textcolor{black}{\cite{huang2026pointworld}}
		& \second{112.5} & \second{49.1} & \second{76.1}
		& 9.22 & 9.38 & 7.55
		& 77.5 & 45.9 & 59.1
		& 105 & 62 & 80
		& 18.1 & 14.4 & 17.4 \\
		\textcolor{black}{+ Geom. correction~\cite{zhang2025pgnd}}
		& 139.2 & 65.2 & 87.4
		& \second{7.34} & 8.92 & \second{6.29}
		& 74.9 & \second{41.1} & \second{56.0}
		& 101 & \second{55} & \second{76}
		& 18.3 & 15.2 & 18.5 \\
		\textcolor{black}{+ Rigid fitting~\cite{kabsch1976rotation}}
		& 129.5 & 57.9 & 83.1
		& 8.18 & \second{7.69} & 6.75
		& \second{72.8} & 42.4 & 56.7
		& \second{98} & 57 & 77
		& \best{0.0} & \best{0.0} & \best{0.0} \\
		\midrule
		\textbf{\ours}
		& \best{18.2} & \best{10.5} & \best{12.1}
		& \best{4.82} & \best{6.08} & \best{4.08}
		& \best{61.8} & \best{35.2} & \best{48.4}
		& \best{75.4} & \best{44.2} & \best{58.6}
		& \second{4.9} & \second{2.9} & \second{3.2} \\
		\bottomrule
	\end{tabular}
\end{table*}

%% file: fig/route_results_table.tex
\begin{table*}[t]
\caption{\textbf{Interaction video prediction on DROID and BEHAVIOR-1K.}
CD denotes Chamfer distance.
Units are px for 2D trajectory error, mm for CD, centroid displacement error,
and background drift, \% for overstretching, and dB for PSNR.}
\label{tab:main}
\label{tab:k4d}
\centering
\color{black}
\fontsize{8.5}{9.8}\selectfont
\setlength{\tabcolsep}{2.2pt}
\renewcommand{\arraystretch}{1.08}
\begin{tabular*}{\textwidth}{@{\extracolsep{\fill}}l*{10}{c}@{}}
\toprule
\multirow{2}{*}{Method}
& \multicolumn{3}{c}{Motion error}
& \multicolumn{3}{c}{Geometric diagnostics}
& \multicolumn{4}{c}{Visual quality} \\
\cmidrule(lr){2-4}\cmidrule(lr){5-7}\cmidrule(l){8-11}
& 2D traj.$\downarrow$ & CD$\downarrow$ & Centroid disp.$\downarrow$
& Overstretching$\downarrow$ & Edge strain$\downarrow$ & BG drift$\downarrow$
& PSNR$\uparrow$ & SSIM$\uparrow$ & LPIPS$\downarrow$ & FVD$\downarrow$ \\
\midrule
Wan2.1-I2V~\cite{wan2025}
& 31.84 & 104.8 & 119.2
& 8.4 & 0.87 & 188.5
& 14.38 & 0.571 & 0.373 & 531 \\
Ctrl-World~\cite{guo2025ctrlworld}
& 12.15 & 70.4 & \second{41.0}
& 7.6 & 0.64 & \second{10.1}
& \second{22.12} & 0.748 & 0.308 & 309 \\
Kinema4D~\cite{kinema4d2026}
& 12.02 & \second{65.8} & 57.6
& \second{3.9} & \second{0.59} & 10.2
& 19.31 & 0.793 & 0.245 & 260 \\
TesserAct~\cite{zhen2025tesseract}
& 26.79 & 92.5 & 104.2
& 12.8 & 0.91 & 103.10
& 16.07 & 0.754 & 0.242 & 435 \\
PointWorld~\cite{huang2026pointworld} + \textcolor{black}{\rwRAVA{}}
& \second{9.58} & 68.1 & 55.7
& 7.2 & 0.65 & 11.3
& 21.95 & \second{0.812} & \second{0.135} & \second{215} \\
\midrule
\textbf{\ours}
& \best{8.12} & \best{61.2} & \best{38.0}
& \best{2.8} & \best{0.50} & \best{8.2}
& \best{23.92} & \best{0.865} & \best{0.093} & \best{149} \\
\bottomrule
\end{tabular*}
\end{table*}

%% file: fig/qualitative_comparison.tex
\begin{figure*}[t]
\centering
\begingroup
\resizebox{\textwidth}{!}{%
\begin{tikzpicture}[x=1bp,y=-1bp]
\node[anchor=north west,inner sep=0] at (0,0)
{\includegraphics[width=508bp]{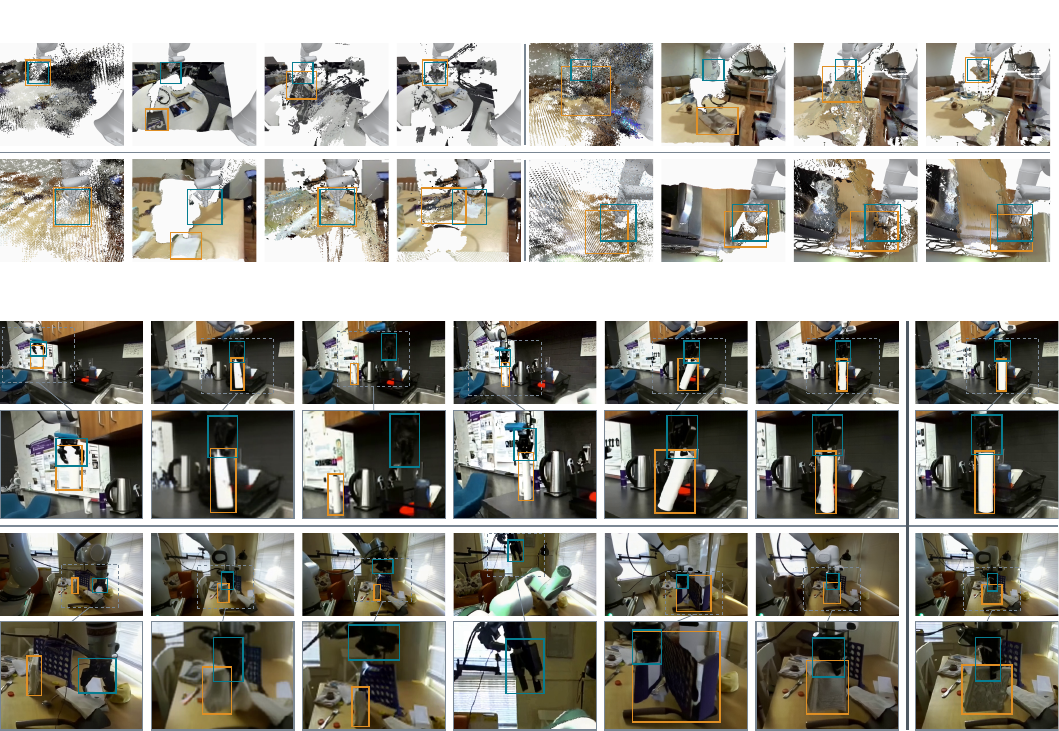}};
		\tikzset{
			comparisonheading/.style={
				anchor=west,inner sep=0,yshift=-2.5bp,
				font=\rmfamily\mdseries\fontsize{8.5}{9}\selectfont},
			comparisonmethod/.style={
				anchor=north,inner sep=0,align=center,yshift=-2.5bp,
				font=\rmfamily\mdseries\fontsize{7}{8}\selectfont},
			comparisonmethodbold/.style={
				comparisonmethod,font=\rmfamily\bfseries\fontsize{7}{8}\selectfont}
		}
		
		\node[comparisonheading] at (0,4.6) {(a) 3D interaction};
		\node[comparisonheading] at (0,137.907292) {(b) Video prediction};
		
		\node[comparisonmethod] at (29.750000,10.000000) {Kinema4D\,\cite{kinema4d2026}};
		\node[comparisonmethod] at (93.250000,10.000000) {TesserAct\,\cite{zhen2025tesseract}};
		\node[comparisonmethod] at (156.750000,10.000000) {PointWorld\,\cite{huang2026pointworld}};
		\node[comparisonmethodbold] at (220.250000,10.000000) {RehearsalWorld};
		\node[comparisonmethod] at (283.750000,10.000000) {Kinema4D\,\cite{kinema4d2026}};
		\node[comparisonmethod] at (347.250000,10.000000) {TesserAct\,\cite{zhen2025tesseract}};
		\node[comparisonmethod] at (410.750000,10.000000) {PointWorld\,\cite{huang2026pointworld}};
		\node[comparisonmethodbold] at (474.250000,10.000000) {RehearsalWorld};
		
		\node[comparisonmethod] at (34.285714,143.307292) {Wan2.1-I2V\,\cite{wan2025}};
		\node[comparisonmethod] at (106.857143,143.307292) {Ctrl-World\,\cite{guo2025ctrlworld}};
		\node[comparisonmethod] at (179.428571,143.307292) {Kinema4D\,\cite{kinema4d2026}};
		\node[comparisonmethod] at (252.000000,143.307292) {TesserAct\,\cite{zhen2025tesseract}};
		\node[comparisonmethod] at (324.571429,143.307292) {PointWorld\,\cite{huang2026pointworld} + \textcolor{black}{\rwRAVA{}}};
		\node[comparisonmethodbold] at (397.142857,143.307292) {RehearsalWorld};
		\node[comparisonmethodbold] at (473.714286,143.307292) {GT};
	\end{tikzpicture}%
}
\endgroup
\caption{\textbf{Qualitative comparison of 3D interactions and video predictions.}
	(a) Four point-cloud comparisons. TesserAct point clouds are back-projected
	from generated depth. Robot geometry is shown consistently across methods.
	(b) Generated and ground-truth (GT) frames for two scenes (rows 1 and 3),
	with corresponding close-ups (rows 2 and 4) at matched magnification
	within each scene.
	Orange boxes mark target objects; cyan boxes mark contact regions in (a)
	and grippers in (b).}
\label{fig:qualitative}
\end{figure*}
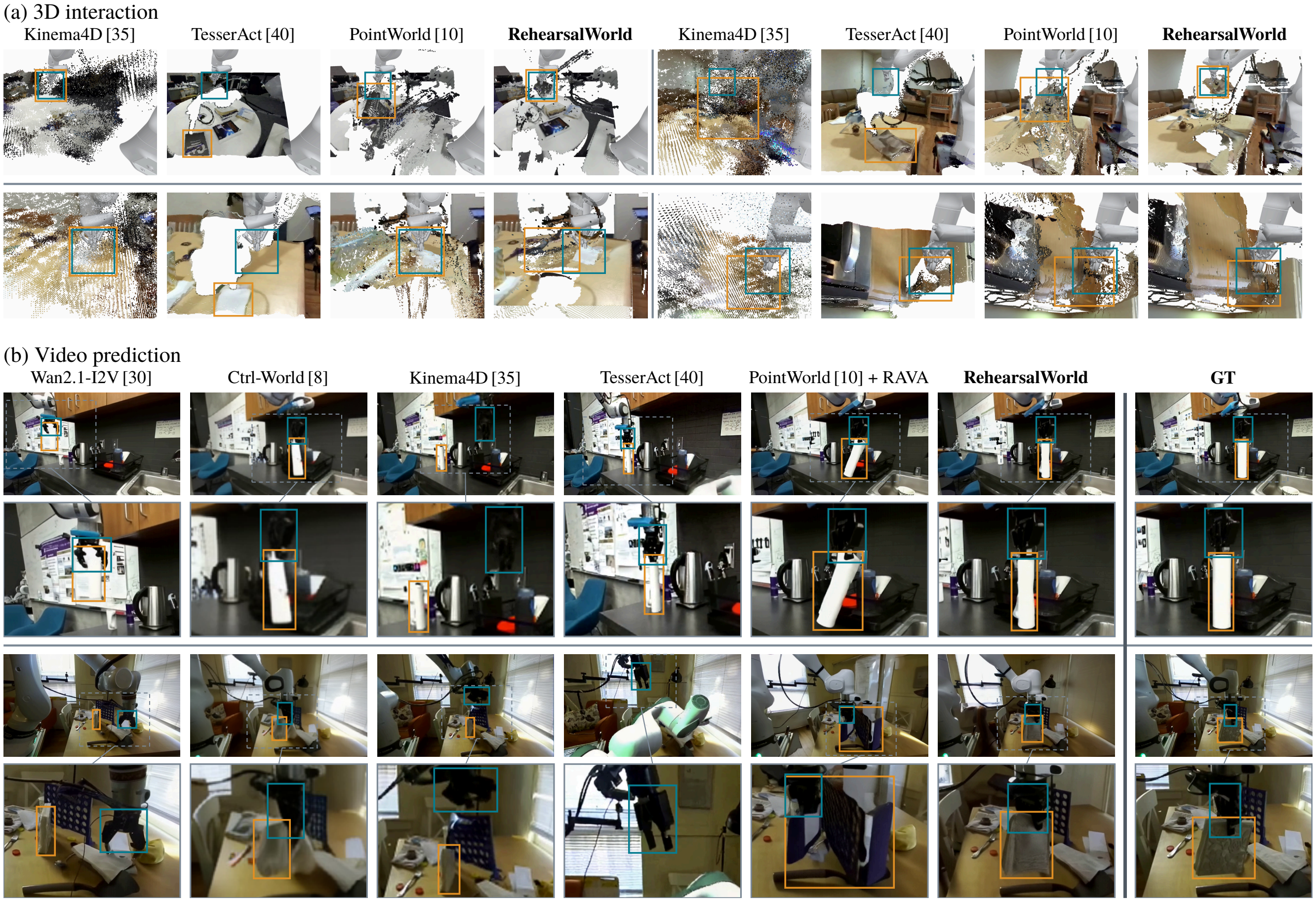

%% file: sec/5_experiments.tex
\section{Experiments}
\label{sec:experiments}
\raggedbottom

{\color{black}
	We evaluate 3D rehearsal on real DROID~\cite{khazatsky2024droid} sequences for grasp preservation, trajectory accuracy, and object structure, then test whether these gains carry over to video generation on DROID and BEHAVIOR-1K~\cite{li2022behavior1k}.
	Ablations isolate the effects of physical constraints and render alignment. We further demonstrate material and viewpoint editing.
}

\subsection{Experimental Setup}
\label{sec:setup}
\label{sec:cfdroid}

{\color{black}
	\noindent \textbf{Data and evaluation protocol.} \enspace
	For 3D evaluation, we randomly sampled 1,200 independent held-out DROID sequences, stratified by object type (500 deformable, 200 articulated, and 500 rigid).
	The DROID training and test splits share neither collection scenes nor interaction actions.
	\textcolor{black}{We trained \rwMMH{} on 15,000 independent DROID~\cite{khazatsky2024droid} sequences with action-response supervision.}
	We combined SAM3 region proposals~\cite{carion2025sam3} and CoWTracker point tracks~\cite{lai2026cowtracker} with depth, camera calibration, and gripper contact to label responsive objects and reliable static points in 3D, excluding robot and uncertain points.
	We fine-tuned \textcolor{black}{\rwRAVA{}} on 3,000 render--RGB video pairs from DROID and BEHAVIOR-1K~\cite{li2022behavior1k}.
	During RAVG training, we used TGP renders of reference 3D trajectories as conditions. RGB videos at the same timestamps and viewpoints served as targets.
	At inference, the same TGP renders predicted 3D trajectories instead. No future observations are used.
}

{\color{black}
\noindent \textbf{Implementation details.} \enspace
\textcolor{black}{We set the prediction horizon to $T = 60$\textcolor{black}{, the video resolution to $480 \times 832$, and the frame rate to 15 fps.} \rwMMH{} was trained independently with focal loss ($\gamma = 2$), using AdamW for 1 epoch with a batch size of 8 and a learning rate of $3 \times 10^{-4}$. For RAVG, we fine-tuned the FixAnything~\cite{vuong2026fixanything} adaptation of Wan2.1~\cite{wan2025}. We used BF16, rank-64 LoRA, and a flow-matching objective, optimized with AdamW at a learning rate of $2 \times 10^{-5}$ for 2400 steps, including 20 warmup steps. PointWorld, the video generator backbone, and the VAE remained frozen throughout training.}
}

{\color{black}
	\noindent \textbf{Baselines.} \enspace
	\textcolor{black}{For 3D evaluation by object category, we compare with PointWorld~\cite{huang2026pointworld} and two of its variants.
		One applies PGND-inspired geometric correction~\cite{zhang2025pgnd}, and the other uses global rigid fitting via Kabsch registration~\cite{kabsch1976rotation}.
		Both variants are detailed in the supplement.}
	For video prediction, we compare with Wan2.1-I2V~\cite{wan2025}, Ctrl-World~\cite{guo2025ctrlworld}, Kinema4D~\cite{kinema4d2026}, and TesserAct~\cite{zhen2025tesseract}.
	PointWorld~\cite{huang2026pointworld} + \textcolor{black}{\rwRAVA{}} feeds unrefined PointWorld predictions through the same TGP and video generator as \ours{} to isolate the effect of 3D refinement on video.
}

\subsection{Main Results}
\label{sec:exp_main}

{\color{black}
	\noindent\textbf{3D interaction.}\enspace
	We evaluate 3D interaction rehearsal on deformable, articulated, and rigid objects in terms of grasp stability, trajectory accuracy, and structural preservation.
	As shown in Tab.~\ref{tab:categories}, \ours reduces gripper-relative offset by approximately 79\%--84\% compared with PointWorld.
	The offset decreases from 112.5 to 18.2\,mm for deformable objects, from 49.1 to 10.5\,mm for articulated objects, and from 76.1 to 12.1\,mm for rigid objects.
	Across all three object categories, \ours achieves the lowest pre-grasp, moving-point, and post-grasp free-point errors.
	The post-grasp free-point errors are 75.4, 44.2, and 58.6\,mm, respectively, showing that the gains extend to points not directly attached to the gripper.
	Overstretching rates fall from 14.4\%--18.1\% for PointWorld to 2.9\%--4.9\% for \ours.
	The persistent grasp reference keeps contact points moving with the gripper, while object-level structural constraints limit excessive stretching without imposing rigid motion on deformable objects.
	
	Fig.~\ref{fig:qualitative}(a) illustrates these improvements in the point clouds.
	Compared with the original PointWorld predictions, \ours reduces point scatter and excessive stretching near the grasp, better preserving the spatial relationship between the object and the gripper.
	Fig.~\ref{fig:phys} further compares per-frame overstretching rates for both methods.
}

{\color{black}
	\noindent\textbf{Video prediction.}\label{sec:application}\enspace
	We evaluate video prediction on DROID and BEHAVIOR-1K in terms of motion error, geometric diagnostics, and visual quality.
	\ours ranks first on all ten metrics aggregated across the two datasets (Tab.~\ref{tab:main}).
	To assess how 3D refinement affects the final videos, we compare with PointWorld + \textcolor{black}{\rwRAVA{}}, which uses the original PointWorld trajectories with the same TGP and video generator as \ours.
	Relative to this baseline, 2D trajectory error decreases from 9.58 to 8.12\,px, a 15\% reduction.
	CD falls from 68.1 to 61.2\,mm, and centroid displacement error falls from 55.7 to 38.0\,mm.
	In 3D reconstructions of the generated RGB videos, the overstretching rate drops from 7.2\% to 2.8\%, a 61\% reduction.
	Edge strain decreases from 0.65 to 0.50, and background drift decreases from 11.3 to 8.2\,mm.
	
	Frame-level fidelity also improves. PSNR rises from 21.95 to 23.92\,dB and SSIM from 0.812 to 0.865, while LPIPS falls from 0.135 to 0.093.
	FVD drops from 215 to 149, indicating an improvement at the video level as well.
	With the projection and generation pipeline held fixed, these results show that refining the 3D trajectories improves both motion accuracy and visual fidelity in the final videos.
	Fig.~\ref{fig:qualitative}(b) compares generated and ground-truth frames.
}

\input{fig/human_subjective_table}

{\color{black}
\noindent\textbf{Human evaluation.}\enspace
Fifty evaluators rated pairs of anonymized videos from 300 independent scenes randomly sampled from the test set.
They rated visual naturalness, temporal stability, contact plausibility, and motion accuracy on a five-point scale.
We compared scores using paired $t$-tests with Holm correction. 
As shown in Tab.~\ref{tab:human_preference}, \ours receives the highest scores for contact plausibility and motion accuracy, significantly outperforming all baselines on both dimensions. Its visual naturalness and temporal stability scores are also close to the best in the table, indicating that the gains in interaction quality are accompanied by strong visual quality.}

\subsection{Ablation Study}
\label{sec:exp_ablation}
\input{fig/combined_ablations_table}

{\color{black}
We conduct ablations on RehearsalWorld. Firstly, we evaluate video prediction quality by removing 3DIR and omitting RAVA fine-tuning. It is shown in Tab. 4 that 3DIR reduces 2D trajectory error and that RAVA fine-tuning improves visual fidelity. The ablations of 3DIR then assess 3D rehearsal quality. We remove RAOG and PGM separately, disable CSR’s structural, interaction, and physical-effect constraints individually and jointly, and remove all 3DIR components. As shown in Tab. 5, the full model achieves lower gripper-relative offset, overstretching rate, and motion error than each ablated variant. These results indicate that the proposed design helps maintain stable grasps, limit excessive stretching, and improve motion prediction accuracy.
}

\subsection{Material and Viewpoint Editing}
\label{sec:exp_id}

\begin{figure}[!t]
	\centering
	\includegraphics[width=\columnwidth]{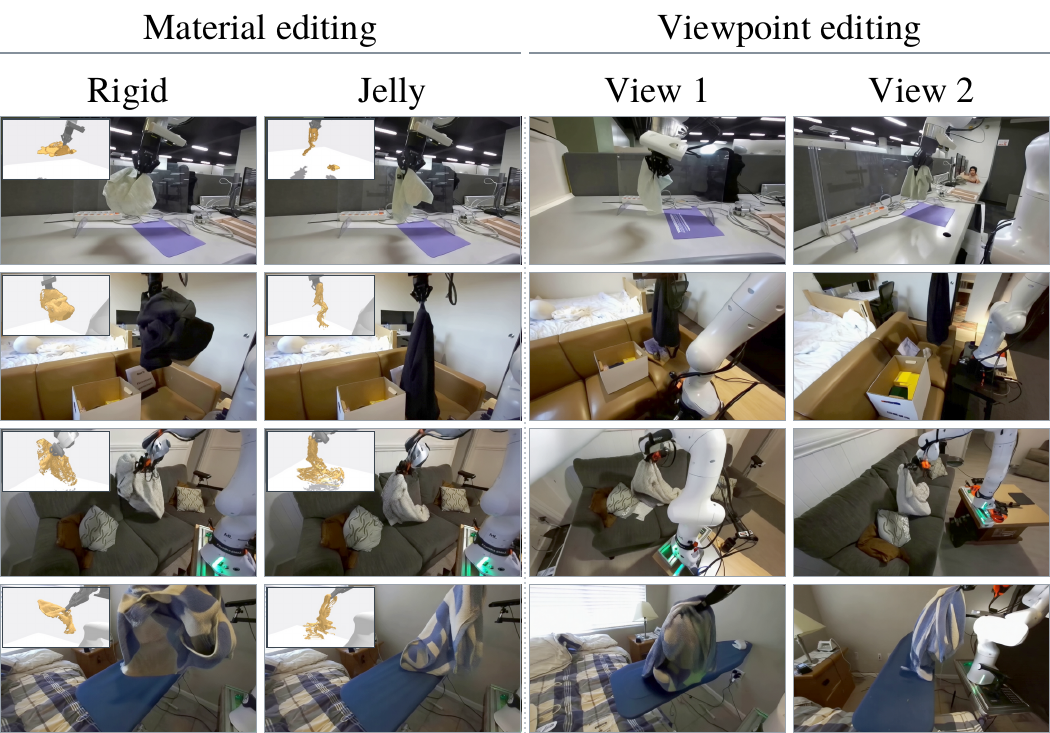}
	\caption{\textbf{Decoupled material and viewpoint editing.}
		Two material settings yield different responses with the initial observation
		and actions fixed. Two views show the same cloth rehearsal.}
	\label{fig:video}
	\label{fig:material_response}
\end{figure}

{\color{black}
	For material editing, we test rigid, jelly, and cloth settings for the target
	object while keeping the initial RGB-D observation and robot actions fixed.
	The three settings produce distinct 3D trajectories and deformation patterns
	(Fig.~\ref{fig:material_tracks}).
	For viewpoint editing, we change only the camera path to show the same
	interaction from different viewpoints, leaving the 3D trajectories and contact
	sequence unchanged.
	Fig.~\ref{fig:video} shows the resulting videos for both types of edits.
	Both reuse TGP and the video generator without retraining.
}

%% file: fig/human_subjective_table.tex
\begin{table}[t]
	\caption{\textbf{Human evaluation on a five-point scale.}
		Higher scores are better. Parentheses below baseline means show
		Holm-adjusted $p$ values for paired comparisons with \ours{}.}
	\label{tab:human_preference}
	\centering
	\small
	\setlength{\tabcolsep}{0.8pt}
	\renewcommand{\arraystretch}{1.08}
	\newcommand{\rwHumanScore}[2]{%
		\shortstack{#1\\{\footnotesize$(#2)$}}}
	
	\begin{tabular*}{\columnwidth}{@{\extracolsep{\fill}}lcccc@{}}
		\toprule
		Method & Visual & Temporal & Contact & Motion \\
		\midrule
		Ctrl-World~\cite{guo2025ctrlworld}
		& \rwHumanScore{3.45}{<0.001}
		& \rwHumanScore{3.51}{<0.001}
		& \rwHumanScore{3.05}{<0.001}
		& \rwHumanScore{3.12}{<0.001} \\
		Kinema4D~\cite{kinema4d2026}
		& \rwHumanScore{\textbf{4.21}}{0.364}
		& \rwHumanScore{4.05}{0.088}
		& \rwHumanScore{3.62}{<0.001}
		& \rwHumanScore{3.54}{<0.001} \\
		PointWorld~\cite{huang2026pointworld} + \textcolor{black}{\rwRAVA{}}
		& \rwHumanScore{4.18}{0.512}
		& \rwHumanScore{\textbf{4.12}}{0.215}
		& \rwHumanScore{3.88}{<0.01}
		& \rwHumanScore{3.79}{<0.001} \\
		\textbf{\ours}
		& 4.15 & 4.09 & \textbf{4.22} & \textbf{4.18} \\
		\bottomrule
	\end{tabular*}
\end{table}

%% file: fig/combined_ablations_table.tex
\begin{table}[t]
\captionsetup{skip=3pt}
\caption{\textbf{Ablation of video prediction.}}
\label{tab:render}
\centering
\color{black}
\fontsize{8}{9.5}\selectfont
\setlength{\tabcolsep}{2.2pt}
\renewcommand{\arraystretch}{1.08}
\newcommand{\rwAblHead}[1]{%
\ensuremath{\vcenter{\hbox{\shortstack[c]{#1}}}}%
}
\begin{tabular}{@{}>{\raggedright\arraybackslash}m{0.32\columnwidth}@{}*{4}{>{\centering\arraybackslash}m{0.17\columnwidth}@{}}}
	\toprule
	\centering\rwAblHead{Configuration}
	& \rwAblHead{PSNR$\uparrow$\\(dB)}
	& \rwAblHead{LPIPS$\downarrow$}
	& \rwAblHead{Object-\\region\\LPIPS$\downarrow$}
	& \rwAblHead{2D\\trajectory\\error (px)$\downarrow$} \\
	\midrule
	\textbf{\ours}
	& \best{23.92} & \best{0.093} & \best{0.112} & \best{8.12} \\
	w/o 3DIR
	& \second{21.95} & \second{0.135} & \second{0.158} & 9.58 \\
	w/o alignment
	& 14.27 & 0.418 & 0.512 & \second{8.25} \\
	w/o both
	& 13.80 & 0.450 & 0.495 & 15.10 \\
	\bottomrule
\end{tabular}

\par\vspace{4pt}
\caption{\textbf{Ablation of 3DIR components.}}
\label{tab:ablation}
\renewcommand{\arraystretch}{1.14}
\begin{tabular}{@{}>{\raggedright\arraybackslash}m{0.32\columnwidth}@{}*{4}{>{\centering\arraybackslash}m{0.17\columnwidth}@{}}}
\toprule
\centering\rwAblHead{Configuration}
& \rwAblHead{Gripper\\Offset\\(mm)$\downarrow$}
& \rwAblHead{Edge\\Strain\\$\downarrow$}
& \rwAblHead{Overstretch\\Rate\\(\%)$\downarrow$}
& \rwAblHead{Motion\\Error\\(mm)$\downarrow$} \\
\midrule
\textbf{\ours}
& \best{13.6} & \second{0.52} & \best{3.7} & \best{48.4} \\
w/o RAOG
& 35.2 & \best{0.45} & 12.5 & 55.1 \\
w/o PGM
& 70.5 & 1.02 & 14.2 & 58.6 \\
\midrule
\multicolumn{5}{@{}l@{}}{\textit{Constraints within CSR}} \\
\hspace{0.8em}w/o structure
& 28.6 & 1.54 & 25.8 & \second{52.3} \\
\hspace{0.8em}w/o interaction
& 75.1 & 0.86 & 18.4 & 56.2 \\
\hspace{0.8em}w/o physical effects
& \second{22.5} & 0.57 & \second{6.8} & 57.2 \\
\hspace{0.8em}w/o CSR constraints
& 74.3 & 1.54 & 15.8 & 59.1 \\
\midrule
w/o all
& 79.2 & 1.71 & 16.6 & 60.8 \\
\bottomrule
\end{tabular}
\end{table}

%% file: sec/7_conclusion.tex
\section{Conclusion}
\label{sec:conclusion}

We present \ours, an editable robot world model that decouples 3D interaction rehearsal from video synthesis.
Experiments show that preserving object structure and grasp attachments in 3D improves motion accuracy and video fidelity.
This design also supports independent editing of material responses and camera viewpoints.

\noindent\textbf{Limitations and future work.}\enspace
Our evaluation focuses primarily on grasping interactions, and we illustrate material editing through qualitative examples.
Future work will extend the framework to a broader range of contact scenarios.

%% file: supplement.tex
\makeatletter
\setlength{\@dblfptop}{0pt}
\setlength{\@dblfpsep}{10pt}
\makeatother

\renewcommand{\dbltopfraction}{0.95}
\renewcommand{\dblfloatpagefraction}{0.75}
\setcounter{dbltopnumber}{4}
\renewcommand{\textfraction}{0.05}
\setlength{\textfloatsep}{10pt plus 2pt minus 2pt}
\setlength{\dbltextfloatsep}{10pt plus 2pt minus 2pt}
\newcommand{\suppheading}[1]{%
  \par\addvspace{6pt}\noindent\textbf{#1}\hspace{0.7em}\ignorespaces}
\clearpage
\setcounter{figure}{0}
\setcounter{table}{0}
\setcounter{equation}{0}
\renewcommand{\theHfigure}{supp.\arabic{figure}}
\renewcommand{\theHtable}{supp.\arabic{table}}
\renewcommand{\theHequation}{supp.\arabic{equation}}
\appendix
\renewcommand{\thesection}{S\arabic{section}}
\renewcommand{\thesubsection}{\thesection.\arabic{subsection}}
\renewcommand{\thefigure}{S\arabic{figure}}
\renewcommand{\thetable}{S\arabic{table}}
\renewcommand{\theequation}{S\arabic{equation}}
\raggedbottom
\twocolumn[{%
  \pdfbookmark[0]{Supplementary Material}{supplementary}
  \begin{center}{\LARGE\bfseries Supplementary Material\par}\end{center}
  \input{fig/supp_opening}
}]

\section{Experimental Setup and Evaluation}
\label{sec:supp_data}

\subsection{Implementation Details}
\label{sec:supp_training_details}
\suppheading{Inputs and outputs.}
\ours uses a pretrained 3D point-flow model~\cite{huang2026pointworld} as its motion prior, given an initial RGB-D observation, camera calibration, and robot actions $q_{0:T}$. Response-aware object grouping (RAOG) identifies groups of responsive points, persistent grasp memory (PGM) maintains grasp references, and the contact structure refiner (CSR) applies the constraints. The resulting trajectories $\widehat Z_{0:T}$ preserve point correspondence.

Temporal geometry projection (TGP) renders the rehearsal along the specified camera path. Render-aligned video adaptation (RAVA) uses these renders to generate the RGB video. We use the observation view by default and also support material and camera-path edits. We predict $T=60$ steps and output the initial frame and 60 subsequent frames at $480\times832$ and 15 fps. The first and last frames are 4.00\,s apart, and playback lasts about 4.07\,s.

We train the RAOG prediction head and RAVA's LoRA weights separately. The point-flow model, video backbone, and VAE stay frozen. Secs.~\ref{sec:supp_method} and~\ref{sec:supp_editing} give the implementation details, along with main-paper Sec.~\ref{sec:method}.

\suppheading{Optimization.}
Tab.~\ref{tab:supp_training} summarizes both training runs. We train RAOG with AdamW for one epoch at batch size 8 and a learning rate of $3\times10^{-4}$. For RAVA, we start from Wan2.1~\cite{wan2025} adapted by FixAnything~\cite{vuong2026fixanything} and update only the LoRA weights, with rank and scale both set to 64. We use BF16 and AdamW for 2,400 steps, including 20 warmup steps, at a learning rate of $2\times10^{-5}$.

\subsection{Training Data}
\label{sec:supp_data_construction}
\input{fig/supp_training_settings}

We train RAOG on DROID~\cite{khazatsky2024droid} and RAVA on both DROID and BEHAVIOR-1K~\cite{li2022behavior1k} (Fig.~\ref{fig:supp_opening}(a)). DROID provides synchronized multiview RGB-D, camera calibration, and robot states for about 76,000 demonstrations in 564 real scenes. BEHAVIOR-1K defines 1,000 everyday activities across 50 interactive scenes in OmniGibson. Tab.~\ref{tab:supp_training} summarizes the training sets and settings.

\suppheading{RAOG training.}
For 15,000 distinct DROID sequences, we pair initial 3D points with motion features extracted by frozen PointWorld from RGB-D and robot actions. To label these points, we align RGB frames with trajectory windows and discard misaligned sequences. SAM3~\cite{carion2025sam3} segments the robot and candidate objects, and CoWTracker~\cite{lai2026cowtracker} tracks the candidates. Relative motion around grasp closure identifies the manipulated objects. We project their masks onto the initial point cloud using depth and camera calibration (Fig.~\ref{fig:supp_training_supervision}, top).

Focal loss ($\gamma=2$) supervises the point responses, with reliable responsive-object points as positives and points that remain still across multiple frames as background negatives. We exclude robot and ambiguous points and discard sequences without a valid response region. For the type loss, we use cross-entropy on groups whose categories map reliably to rigid, deformable, or articulated. RAOG supplies responsive points, object-point groups, and types to PGM and CSR. Later video frames are used only to construct offline labels.

\suppheading{RAVA training.}
Of the 3,000 render--RGB pairs, 2,700 have recorded DROID targets and 300 have simulated BEHAVIOR-1K targets. Starting from the initial RGB-D scene, TGP renders reference 3D trajectories at the target frame times and viewpoints, aligning object motion and robot pose across each pair. The coarse renders provide motion and occlusion cues even where appearance is incomplete (Fig.~\ref{fig:supp_training_supervision}, bottom).

\input{fig/supp_training_supervision}

The initial RGB frame provides the appearance reference. The frozen VAE encodes the coarse renders as temporal conditioning, and a flow-matching loss on the paired RGB video trains the LoRA weights. Subsequent target frames supply supervision only. At inference, we use the same projection and conditioning pipeline with rehearsed trajectories in place of reference trajectories.

\subsection{Evaluation Metrics}
\label{sec:supp_metrics}\label{sec:supp_native_metrics}\label{sec:supp_protocols}
We evaluate native 3D trajectories on 1,200 held-out DROID sequences: 500 deformable, 200 articulated, and 500 rigid. The training and test splits share neither collection scenes nor interaction actions. We evaluate video prediction on DROID and BEHAVIOR-1K and score native trajectories separately from geometry reconstructed from RGB. Main-paper Sec.~4.1 describes the splits and baselines, and Sec.~4.2 reports the results.

\suppheading{Baselines.}
Both 3D variants in main-paper Tab.~1 start from PointWorld. The geometric refinement variant anchors the background and applies bounded rigid fitting or PGND-inspired particle--grid displacement smoothing~\cite{zhang2025pgnd}, depending on object type. It then adjusts clearance from the robot surface and uses the corrected endpoint for the next prediction. The global rigid fit uses Kabsch registration~\cite{kabsch1976rotation} at each frame to fit one rotation and translation from the initial object-point group to the original prediction. In main-paper Tab.~2, PointWorld + RAVA renders the original PointWorld trajectories using the same TGP and video generator as \ours. This comparison tests how 3D trajectory refinement affects the generated video.

\suppheading{Native 3D trajectories.}
We compare rehearsed points with their reference correspondences. The reference fixes the grasp timing, moving points, and contact points, so all methods are evaluated on the same point sets and time windows.
\begin{itemize}
\setlength{\itemsep}{0pt}\setlength{\parsep}{0pt}\setlength{\topsep}{2pt}
\item \textbf{3D position errors.} Pre-grasp error, moving-point error, post-grasp free-point error, and ablation motion error average 3D position distances over their respective point sets and time windows, in mm. The reference defines post-grasp free points independently of PGM's predicted contact set.
\item \textbf{Gripper-relative offset.} We use object points within 4\,cm of the gripper center at closure in the reference. At each time, we center the predicted point positions at the gripper and retain the world axes. We measure the Euclidean change from the relative position at closure and average over evaluated points and times, in mm. The evaluated set is independent of PGM's predicted attached points.
\item \textbf{Structural deformation.} Edge strain averages $|L_t/L_0-1|$, where $L_0,L_t$ are the initial and current lengths of each evaluated edge. The overstretch rate (\%) is the fraction of point pairs beyond the stretch threshold. Each experiment fixes its pairs, threshold, and time window across variants.
\end{itemize}

\suppheading{Material solver metrics.}
Tab.~\ref{tab:supp_physical_ablation} reports motion error (mm), particle deformation, and support-plane penetration. Cloth edge strain averages $|L_t/L_0-1|$ over six-neighbor edges initially shorter than 2.5 times the particle spacing. For cloth and jelly, overstretch is the fraction of edges longer than $2L_0$ in the last evaluated frame. Support-plane penetration is the fraction of cloth particle--time pairs more than 1\,cm below the initial support plane.

\suppheading{RGB video motion and reconstructed geometry.}
\begin{itemize}
\setlength{\itemsep}{0pt}\setlength{\parsep}{0pt}\setlength{\topsep}{2pt}
\item \textbf{2D trajectories.} CoTracker3~\cite{karaev2025cotracker3} tracks the same initial object pixels for every method. We compute the 2D trajectory error (px) over jointly visible points.
\item \textbf{Reconstructed geometry.} SpatialTrackerV2's VGGT4Track front end~\cite{xiao2025spatialtrackerv2} reconstructs geometry from RGB. We fit the metric map to initial depth and hold it fixed, without registration to later depth. We compute object-point Chamfer distance, centroid displacement error, and background drift (mm), and evaluate overstretch and strain on a fixed edge set. The background mask is shared across methods and frames.
\end{itemize}

\suppheading{Visual quality.}
We exclude the conditioning frame from fidelity evaluation.
\begin{itemize}
\setlength{\itemsep}{0pt}\setlength{\parsep}{0pt}\setlength{\topsep}{2pt}
\item \textbf{Frame fidelity.} PSNR (dB), SSIM, AlexNet LPIPS, and object-region LPIPS use paired generated and reference frames. For object-region LPIPS, we extract features from complete frames and weight each layer's distance map with the shared reference mask. We exclude empty masks and average frame scores within each sequence before aggregating valid sequences.
\item \textbf{Video distribution.} FVD compares the distributions of I3D spatiotemporal features extracted from equal-length clips.
\end{itemize}

\suppheading{Human evaluation.}
We randomly selected 300 independent scenes from the test set. Fifty evaluators rated anonymized video pairs for visual naturalness, temporal stability, contact plausibility, and motion accuracy. We matched resolution, playback speed, and duration and randomized left--right order. On the five-point scale, 1 indicates severe defects, 3 intermittent visible defects, and 5 no readily visible defects. We average valid scores within each scene and run paired $t$-tests across scenes with Holm correction.
\section{Method Details}
\label{sec:supp_method}
\label{sec:supp_3dir}
\label{sec:supp_state_implementation}
\label{sec:supp_3dir_ablation_definitions}
\subsection{Overview}
Interaction-consistent rehearsal (ICR) refines point-flow predictions through RAOG, PGM, and CSR. Here we specify how the three components group objects, maintain grasp references, and project the constraints in the state update from main-paper Sec.~3.2.

\subsection{Response-Aware Object Grouping (RAOG)}
\label{sec:supp_raog}
RAOG identifies responsive points, groups them into objects, predicts their physical types, and anchors the static background (main-paper Sec.~3.2). It operates on the $N$ initial non-robot scene points. We obtain robot geometry separately from its URDF.

{\color{black}We pool motion features over the $N$ scene points to obtain the mean $g_{\rm avg}=N^{-1}\sum_{j=1}^{N}\mathbf f^j$ and channelwise maximum $g_{\rm max}=\max_{1\leq j\leq N}\mathbf f^j$. At each point, we encode the initial position as $a_j=\tanh(W\rho(z_0^j)+b)$ and compute the response probability:
\begin{equation}
\begin{aligned}
\hat{\mathbf f}^j&=[\mathbf f^j;g_{\rm avg}\odot e^{a_j};
g_{\rm max}\odot e^{a_j}],\\
p_j&=\sigma(g_{\rm resp}(\hat{\mathbf f}^j)),\\
\mathcal P&=\{j:p_j\geq\tau_r\},\\
B&=\{1,\ldots,N\}\setminus\mathcal P
\end{aligned}
\label{eq:supp_raog_response}
\end{equation}
The pointwise MLP $g_{\rm resp}$ produces the response logit. The exponential and products operate channelwise.}

We group responsive points by proximity in the initial scene and similarity of their predicted displacement~\cite{song2024geometryconsistency}. {\color{black}Focal loss supervises point responses. We map SAM3 categories to the rigid, deformable, and articulated classes and apply the type cross-entropy to groups with reliable matches. Ambiguous groups do not contribute to this loss. Type supervision follows the category mapping independently of motion magnitude.}

{\color{black}Writing the predicted displacement as $d_t^j=z_t^j-z_0^j$, we connect points among the 16 nearest neighbors in the initial scene when both tests hold:
\begin{equation}
\begin{gathered}
\|z_0^j-z_0^k\|\leq r_{\rm pos},\\
\Bigl(\tfrac{1}{T}\sum\nolimits_{t=1}^{T}\|d_t^j-d_t^k\|^2\Bigr)^{1/2}
\leq r_{\rm mot}.
\end{gathered}
\label{eq:supp_raog_grouping}
\end{equation}
The connected components define the object-point groups $\mathcal P^i$. For each group, we concatenate the mean and maximum of the original features and take the type MLP's most probable class as $\kappa^i$.}

RAOG returns the object set $\mathcal O=\{(\mathcal P^i,\kappa^i)\}_{i=1}^{M}$ and the static background $B$. Group membership stays fixed over the rollout, and $\hat z_t^j=z_0^j\ (j\in B)$ anchors the background at the observed positions. The predicted type $\kappa^i$ selects the default structural constraints. Users set the material-editing parameters independently. {\color{black}Grouping uses no future reference trajectories.}

\subsection{Persistent Grasp Memory (PGM)}
\label{sec:supp_pgm}
PGM stores attached points, gripper-frame positions, and reference edge lengths at grasp onset. The references persist through the grasp and refresh when the object is grasped again (main-paper Sec.~3.2).

{\color{black}Let $c_t^g\in\{0,1\}$ indicate gripper closure from action $q_t$, and let $G_t^{\rm grip}$ be the sampled gripper surface. We initialize $b_0^i=0$ and update the attachment state for $t\geq1$ as follows:
\begin{equation}
\begin{aligned}
d_t^i&=\min_{j\in\mathcal P^i,\,g\in G_t^{\rm grip}}
\|z_t^j-g\|,\\
b_t^i&=c_t^g\,\mathbb 1\{b_{t-1}^i=1\ \text{or}\ d_t^i\leq r_g\}.
\end{aligned}
\label{eq:supp_pgm_trigger}
\end{equation}
The onset time $t_c^i$ is the latest transition in $b_t^i$ from 0 to 1. At onset, we select the closest eligible points, up to $n_a$, within $r_a$ of the gripper to form $\mathcal A^i$ and update the grasp reference. We use $r_g=0.02\,\mathrm m$, $r_a=0.03\,\mathrm m$, and $n_a=16$. Opening the gripper sets $b_t^i=0$. PGM retains the structural reference after release and stores a new contact set and reference at the next grasp.}

At grasp onset, we store the gripper-frame positions and reference edge lengths:
\begin{equation}
r_j^i=(T_{t_c^i}^{g})^{-1}z_{t_c^i}^{j},\qquad
\ell_{jk}^{c,i}=\|z_{t_c^i}^{j}-z_{t_c^i}^{k}\|,
\label{eq:supp_pgm_reference}
\end{equation}
Here $j\in\mathcal P^i$ and $(j,k)\in\mathcal E^i$, where $\mathcal E^i$ is the graph built at grasp onset within the object-point group. {\color{black}Before the first grasp, we build this graph from the initial geometry and set $\ell_{jk}^{c,i}=\|z_0^j-z_0^k\|$. At grasp onset, we update the references with Eq.~\eqref{eq:supp_pgm_reference}. We connect each point to its $k_n=8$ nearest neighbors within the group and symmetrize the graph.}

The current gripper pose maps the stored positions to contact targets $z_{t,\mathrm{target}}^j=T_t^g r_j^i$ for $j\in\mathcal A^i$~\cite{longhini2025clothsplatting}. Free points respond to the attached points through internal structural constraints. Storing the whole group therefore still allows deformation.

{\color{black}Keeping the contact set and local positions fixed during a grasp limits drift from per-frame re-estimation. We use the initial geometry before the first grasp and deactivate the targets on release.}

\subsection{Contact Structure Refiner (CSR)}
\label{sec:supp_csr}
CSR refines the candidate object state $Z_t^i$ by projecting the three constraint groups in main-paper Sec.~3.2. The physical type $\kappa^i$ selects the configuration $\mathcal Q_{\kappa^i}$, and PGM supplies the targets from $\mathcal H_t^i$. {\color{black}For iterate $X^{(m)}$, we define the relaxed projection $\widetilde P_\zeta(X)=X+\eta_\zeta(P_\zeta(X)-X)$ for $\zeta\in\{\mathrm{phys},\mathrm{in},\mathrm{out}\}$. The update is
\begin{equation}
\begin{aligned}
X^{(0)}&=\Pi_{t,\mathrm g}^i(Z_t^i),\\
X^{(m+1)}&=(\Pi_{t,\mathrm g}^i\circ\widetilde P_{\rm out}
\circ\widetilde P_{\rm in}\circ\widetilde P_{\rm phys})(X^{(m)}).
\end{aligned}
\label{eq:supp_csr_solver}
\end{equation}}
Each sweep applies physical effects, internal structure, and environmental interaction in order. The final projection $\Pi_{t,\mathrm g}^i$ restores active contacts to their gripper targets.

\begin{itemize}[leftmargin=1em,labelsep=0.4em,topsep=2pt,itemsep=2pt,parsep=0pt]
\item \textbf{Internal structural constraints.}\enspace
The internal projection acts on each group independently. Its edges lie within the object-point group.

\begin{itemize}[leftmargin=1em,labelsep=0.4em,topsep=2pt,itemsep=2pt,parsep=0pt,label=$\circ$]
\item \textbf{Rigid objects.}\enspace
{\color{black}The rigid projection fits a single transform to the group. The reference $z_{\rm ref}^j$ is the initial position before the first grasp and the position saved by PGM afterward. Given the current positions $x^j$, we solve
\begin{equation}
\begin{aligned}
(R^*,p^*)&=\mathop{\arg\min}_{R\in SO(3),\,p}
\sum_{j\in\mathcal P^i}\|Rz_{\rm ref}^j+p-x^j\|^2,\\
P_{\rm in}(X)_j&=R^*z_{\rm ref}^j+p^*.
\end{aligned}
\label{eq:supp_csr_rigid}
\end{equation}
}The fit gives the group a rigid target. Reapplying the contact targets can move the attached points again, so the final state satisfies both constraints approximately.

\item \textbf{Deformable objects.}\enspace
The deformable projection shortens edges that exceed their allowed length. It uses initial reference lengths before the first grasp and the lengths stored by PGM afterward.
{\color{black}For $(j,k)\in\mathcal E^i$, define the length cap $L_{jk}=(1+\epsilon_{\kappa^i})\ell_{jk}^{c,i}$, excess $e_{jk}=[\|x^j-x^k\|-L_{jk}]_+$, and direction $u_{jk}=(x^j-x^k)/(\|x^j-x^k\|+\varepsilon_{\rm num})$. If $e_{jk}>0$, the free-endpoint weights $w_j=1-\mathbb 1[b_t^i=1,\ j\in\mathcal A^i]$ and $w_k=1-\mathbb 1[b_t^i=1,\ k\in\mathcal A^i]$ allocate the correction:
\begin{equation}
\begin{aligned}
x^j&\leftarrow x^j-\frac{w_j}{w_j+w_k}e_{jk}u_{jk},\\
x^k&\leftarrow x^k+\frac{w_k}{w_j+w_k}e_{jk}u_{jk}.
\end{aligned}
\label{eq:supp_csr_edge}
\end{equation}
We apply the update when $w_j+w_k>0$, average the concurrent edge corrections, and cap their displacement at $c_{\rm in}$. For deformable groups, we use $\epsilon_{\kappa^i}=0.10$, $c_{\rm in}=0.01\,\mathrm m$, and $\varepsilon_{\rm num}=10^{-8}$.} Attached points stay at the gripper targets, and neighboring free points absorb the corrections. The one-sided bound limits excessive stretching and allows bending and folding.

\item \textbf{Articulated objects.}\enspace
Articulated groups use the same graph construction, reference update, and projection in Eq.~\eqref{eq:supp_csr_edge}, with the smaller tolerance {\color{black}$\epsilon_{\kappa^i}=0.05$}. The projection limits local separation and lets neighboring regions rotate relative to each other. This point-based approximation leaves links, joint axes, and angle limits unspecified.
\end{itemize}

\item \textbf{Environmental interaction constraints.}\enspace
{\color{black}The clearance projection limits robot penetration at points outside the active contact set. Let $s_t(x^j)$ be the approximate signed distance to the sampled URDF surface (negative inside) and $n_t(x^j)$ its outward normal. We apply
\begin{equation}
x^j\leftarrow x^j+
\min\{c_{\rm clear},[d_{\rm clear}-s_t(x^j)]_+\}
n_t(x^j).
\label{eq:supp_csr_clearance}
\end{equation}
We restore active contacts to their gripper targets with
\begin{equation}
\Pi_{t,\mathrm g}^i(X)_j=
\begin{cases}
T_t^g r_j^i,&b_t^i=1,\ j\in\mathcal A^i,\\
x^j,&\text{otherwise}
\end{cases}
\label{eq:supp_csr_operators}
\end{equation}
We use $d_{\rm clear}=0.003\,\mathrm m$ and $c_{\rm clear}=0.002\,\mathrm m$.} Both projections use the current robot pose to maintain clearance and grasp attachment.

\item \textbf{Physical-effect constraints.}\enspace
{\color{black}We approximate gravity with a downward correction for free points and support friction with tangential damping:
\begin{equation}
\begin{aligned}
h_{\min,j}^i&=z_{t_{\rm ref}^i,z}^j-d_{\rm floor},\\
h_j^+&=\max\{h_{\min,j}^i,h_j-\Delta h_0\rho^m\},\\
x_{\parallel}^{j,+}&=\hat z_{t-1,\parallel}^j
+(1-\mu)(x_{\parallel}^j-\hat z_{t-1,\parallel}^j).
\end{aligned}
\label{eq:supp_csr_physical}
\end{equation}
We apply the downward step only to free points satisfying $x_z^j-z_{t_{\rm ref}^i,z}^j>h_{\rm lift}$, and damp tangent motion only at support contacts detected in the static scene. The reference time $t_{\rm ref}^i$ is 0 before the first grasp and $t_c^i$ afterward. We use $h_{\rm lift}=d_{\rm floor}=0.01\,\mathrm m$, $\Delta h_0=0.002\,\mathrm m$, $\rho=0.8$, and $\mu=0.3$.} The height bound limits the downward correction.
\end{itemize}

{\color{black}We stop when $\max_j\|X_j^{(m+1)}-X_j^{(m)}\|<\varepsilon_{\rm stop}$ or after $L_{\kappa^i}$ sweeps, then set $\hat Z_t^i=X^{(K_t^i)}$ for the completed sweep count $K_t^i$.

We use $\eta_{\rm in}=0.5$, $\eta_{\rm out}=\eta_{\rm phys}=1$, and $\varepsilon_{\rm stop}=10^{-4}$, with at most 8 sweeps for rigid groups and 12 for deformable and articulated groups. Let $Y_1^{(m)}=\widetilde P_{\rm phys}(X^{(m)})$ and $Y_2^{(m)}=\widetilde P_{\rm in}(Y_1^{(m)})$. The three increments in the main-paper update are
\begin{equation}
\begin{aligned}
\Delta Z_{t,\mathrm{phys}}^i&=\sum_{m=0}^{K_t^i-1}(Y_1^{(m)}-X^{(m)}),\\
\Delta Z_{t,\mathrm{in}}^i&=\sum_{m=0}^{K_t^i-1}(Y_2^{(m)}-Y_1^{(m)}),\\
\Delta Z_{t,\mathrm{out}}^i&=(X^{(0)}-Z_t^i)\\
&\quad{}+\sum_{m=0}^{K_t^i-1}(X^{(m+1)}-Y_2^{(m)}).
\end{aligned}
\label{eq:supp_csr_updates}
\end{equation}
The environmental term includes the initial contact projection, robot clearance, and gripper-target restoration at each sweep. The increments sum to $\hat Z_t^i-Z_t^i$.} Fig.~\ref{fig:supp_state_detail} shows how rehearsal changes local stretch in the basket example.

\input{fig/supp_structure_detail}
\FloatBarrier

\section{Material and Viewpoint Editing}
\label{sec:supp_editing}
A material edit resimulates the target object. A viewpoint edit reprojects the same rehearsal along a new camera path. The particle update and pose interpolation below follow the editing interfaces in main-paper Sec.~3.4. Main-paper Sec.~4.4 reports the editing results.
\subsection{Material-Conditioned Dynamics and Editing}
\label{sec:supp_mcd}
Material-conditioned dynamics (MCD) combines rehearsal trajectories, grasp references, and material controls in one particle update. We vary the target object's parameters $\theta^i$ and hold the initial state, robot actions, grasp targets, and camera fixed.
Fig.~\ref{fig:supp_trajectories} compares generated appearance and surface-point trajectories under the default, rigid, jelly, and cloth settings.

\input{fig/supp_material_trajectories}

\suppheading{Particle-state update.}
We sample dense particles on the observed surface and store their positions and velocities in $S^i=(Y^i,V^i)$, initialized with $S_0^i=(Y_0^i,0)$. The actions determine the URDF robot geometry $G_t$. {\color{black}For each particle $a$, we fix a set of local driving points $\mathcal B_a^i\subseteq\mathcal P^i$ and normalized weights $\nu_{aj}$ at initialization:
\begin{equation}
y_{0,a}^i=\sum_{j\in\mathcal B_a^i}\nu_{aj}z_0^j,
\qquad
\bar x_{t,a}^i=\sum_{j\in\mathcal B_a^i}\nu_{aj}\hat z_t^j.
\label{eq:supp_material_guidance}
\end{equation}
At each PGM grasp onset, these correspondences transfer sparse contacts and gripper-frame references to the particles:
\begin{equation}
\begin{aligned}
\omega_a&=\sum_{j\in\mathcal B_a^i\cap\mathcal A^i}\nu_{aj},\\
\mathcal A^{\mathrm{mat},i}&=\{a:\omega_a>0\},\\
r_a^{\mathrm{mat},i}&=\omega_a^{-1}
\sum_{j\in\mathcal B_a^i\cap\mathcal A^i}\nu_{aj}r_j^i,\\
y_{t,\mathrm{target}}^a&=T_t^g r_a^{\mathrm{mat},i}
\quad(a\in\mathcal A^{\mathrm{mat},i}).
\end{aligned}
\label{eq:supp_material_contact}
\end{equation}
The corrected rehearsal supplies weak trajectory guidance $\bar X^i=(\bar x_{t,a}^i)$. After resimulation, TGP transfers the motion of $Y_t^i$ to the target surface by local weighted skinning (main-paper Sec.~3.3).} Each substep predicts positions from velocity and gravity, projects the constraints, and updates velocities from the corrected displacements:
{\color{black}
\begin{equation}
\begin{aligned}
\widetilde y_a&=y_a+h(v_a+hg),\\
Y^{i,+}&=\mathcal P_{\theta^i,\mathcal A^{\mathrm{mat},i},G_t}
(\widetilde Y^i;\bar X^i),\\
v_a^+&=(1-d)\frac{y_a^+-y_a}{h}.
\end{aligned}
\label{eq:supp_mcd_update}
\end{equation}}
Here $h$ is the substep length, $g$ is gravity, and $d$ controls velocity damping. {\color{black}The projection handles robot collisions, environmental support, grasp targets, and weak trajectory guidance for every material setting.}

\input{fig/controlled_physical_ablation}

\suppheading{Constraints and material responses.}
We collect global and local shape recovery, resistance to edge stretch, velocity damping, and long-range attachment slack in $\theta^i=(a_{\rm g},a_{\rm l},k_{\rm e},d,\delta_{\rm a})$. These controls specify the edit, and the RAOG type $\kappa^i$ selects the default rehearsal constraints.

{\color{black}Rigid and jelly responses use the same shape-matching projection at different spatial scales. We fit the initial shape to the current particle set $\mathcal U$ and project toward the fitted positions:
\begin{equation}
\begin{aligned}
(R_{\mathcal U},q_{\mathcal U})
&=\arg\min_{R\in SO(3),q}\sum_{a\in\mathcal U}
\|R y_{0,a}^i+q-y_a^i\|^2,\\
y_a^i&\leftarrow y_a^i+\alpha_{\mathcal U}
(R_{\mathcal U}y_{0,a}^i+q_{\mathcal U}-y_a^i).
\end{aligned}
\label{eq:supp_mcd_shape_fit}
\end{equation}
For the rigid setting, $\mathcal U$ spans the whole object. The fit favors global shape preservation while allowing translation and rotation. For jelly, we fit overlapping neighborhoods separately and average their corrections. Nearby regions can then move relative to one another, with $d$ controlling how quickly the motion decays. We convert the per-frame weight $a_{\mathcal U}$ to a per-projection weight $\alpha_{\mathcal U}=1-(1-a_{\mathcal U})^{1/(n_s n_I)}$, with $n_s$ substeps per frame and $n_I$ projections per substep.

For cloth, we omit global shape recovery. An edge constraint limits excessive stretching, and long-range attachments limit separation from the grasp region. The surface can still fold and drape:
\begin{equation}
\begin{aligned}
C_{ab}^{\rm edge}(Y)&=\bigl[\|y_a-y_b\|-\ell_{ab}^0\bigr]_+,\\
C_{ab}^{\rm LRA}(Y)&=
\bigl[\|y_a-y_b\|-(1+\delta_{\rm a})d_{ab}^0\bigr]_+.
\end{aligned}
\label{eq:supp_mcd_cloth_constraints}
\end{equation}
Here $\ell_{ab}^0=\|y_{0,a}-y_{0,b}\|$ is the initial edge length and $d_{ab}^0$ the stored initial attachment distance. The controlled ablation in Tab.~\ref{tab:supp_physical_ablation} uses the shortest path in the initial edge graph for that distance. These parameters control the material response without requiring calibrated material constants.
}

\suppheading{Continuous material control and video generation.}
We vary the five controls jointly with $s\in[0,1]$ to move continuously from a cloth-like to a rigid-like response:
\begin{equation}
\begin{aligned}
\theta^i(s)&=\bigl(s^3,\ s,\ 0.95(1-s),\\
&\qquad 0.02+0.03s,\ 0.02+0.08s\bigr).
\end{aligned}
\label{eq:supp_material_control}
\end{equation}
The path runs from the cloth-like response at $s=0$ to the rigid-like response at $s=1$. The locally compliant jelly setting lies off this path. Here $s$ controls the solver and has no calibrated stiffness interpretation. The separate weight $w$ controls trajectory guidance (Sec.~\ref{sec:supp_ablation}). {\color{black}We replace only the target object's trajectory after resimulation. TGP renders the edited trajectory with the same scene, robot, and camera, and the shared generator produces RGB video without retraining.}

\subsection{Viewpoint Editing}
\label{sec:supp_viewpoint}
Viewpoint editing keeps the 3D trajectories and contact sequence fixed (main-paper Sec.~3.4). We interpolate the camera from the observed pose to the target pose.
{\color{black}Let $c_t$ be the world-space camera center and $Q_t$ the camera-to-world rotation. Given the initial pose $(Q_0,c_0)$ and target pose $(Q_\star,c_\star)$, we choose a monotone interpolation coefficient with $\alpha_0=0,\alpha_T=1$:
\begin{equation}
\begin{aligned}
c_t&=(1-\alpha_t)c_0+\alpha_t c_\star,\\
Q_t&=Q_0\operatorname{Exp}\!\left[\alpha_t\operatorname{Log}(Q_0^\top Q_\star)\right].
\end{aligned}
\label{eq:supp_camera_path}
\end{equation}
The relative rotation follows the shortest arc in $SO(3)$. With fixed intrinsics, the world-to-camera extrinsic is $[Q_t^\top\mid -Q_t^\top c_t]$.}
\section{Ablation Analysis}
\label{sec:supp_ablation}
\label{sec:supp_ablations}
We ablate the system at three levels: generated video, native 3D rehearsal, and the edited particle trajectories. Main-paper Tab.~4 varies the trajectories used to render the conditioning video and tests whether the generator benefits from RAVA fine-tuning. Tab.~5 evaluates 3D trajectories directly to isolate how grouping, grasp references, and constraint corrections affect contact, deformation, and motion. Supplementary Tab.~\ref{tab:supp_physical_ablation} holds the guidance trajectory fixed and examines how closely the dense particles follow it under different material constraints. Sec.~\ref{sec:supp_metrics} defines the metrics at each level.

\begin{figure*}[!t]
\centering
\input{fig/supp_selected_pointcloud}
\end{figure*}

\subsection{Material Constraints and Guidance}
\label{sec:supp_physical_controls}
All variants start from the same particles, predicted trajectories, and robot actions on a curated subset of 120 high-precision deformable-object sequences. Variants with grasp attachment use the same grasp targets. We evaluate up to the end of the first annotated grasp, excluding the end frame. The solver runs 40 substeps per frame and three projection iterations per substep. In Tab.~\ref{tab:supp_physical_ablation}, we fix the guidance weight at $w=0.003$ and remove grasp, structure, or support constraints in turn. For the guidance comparison, we keep the physical constraints and vary $w$ to measure changes in motion error, overstretch, and penetration.

\subsection{Discussion}
\label{sec:supp_ablation_discussion}
\suppheading{Motion and appearance in video.}
3D rehearsal primarily improves motion accuracy, and RAVA fine-tuning improves appearance fidelity (main-paper Tab.~4). With RAVA fixed, removing 3DIR raises 2D trajectory error from 8.12 to 9.58\,px. Without RAVA fine-tuning, the 3DIR variant comes close to the full model on trajectory error (8.25\,px), but PSNR falls from 23.92 to 14.27\,dB and object-region LPIPS rises from 0.112 to 0.512. Removing both raises trajectory error to 15.10\,px. The full model combines both benefits and leads on all four metrics.

\suppheading{Contact and structure in 3D rehearsal.}
The full rehearsal reduces gripper-relative offset, overstretch, and motion error (main-paper Tab.~5). RAOG determines which points share a constraint. Without it, mean edge strain drops from 0.52 to 0.45, yet overstretch rises from 3.7\% to 12.5\%. More edges cross the overstretch threshold despite the lower mean strain. Removing PGM raises gripper-relative offset from 13.6 to 70.5\,mm, and removing the interaction constraints raises it to 75.1\,mm. Removing structural constraints produces the highest overstretch rate, 25.8\%. Disabling the physical-effect constraints raises motion error from 48.4 to 57.2\,mm, reflecting the combined contribution of the gravity, support, and friction approximations for free points.

\suppheading{Deformation and trajectory tracking in material edits.}
Attachment and guidance keep the particle motion close to the reference. Structural and support constraints limit overstretch and penetration (Tab.~\ref{tab:supp_physical_ablation}). Removing attachment raises motion error from 12.5 to 118.6\,mm, and removing guidance raises it to 192.4\,mm. Without structural constraints, cloth overstretch rises from 0.89\% to 28.43\%. Without support collision and friction, penetration rises from 0.07\% to 0.67\%. At the strongest guidance setting ($w=0.3$), motion error drops to 10.2\,mm and cloth overstretch rises to 9.06\%. The default setting balances trajectory tracking against the need to limit excessive deformation.

Grouping, grasp references, and constraint projections stabilize 3D contact and structure before render alignment carries the motion into video.

\section{Additional Experimental Results}
\label{sec:supp_results}
Main-paper Sec.~4.2 evaluates native trajectories across three object categories (Tab.~1) and measures motion, reconstructed geometry, and fidelity in DROID and BEHAVIOR-1K videos (Tab.~2). Fig.~5 compares contact-region point clouds and pairs generated frames with references. Fig.~6 tracks overstretch over time. We show which errors rehearsal reduces in additional scenes and extend the editing results in Sec.~4.4 through comparisons under the same robot action.

\subsection{3D Interaction Across Scenes}
Main-paper Tab.~1 reports lower gripper-relative offset and lower post-grasp free-point error for \ours across all three object categories. The offset measures how closely points near the grasp follow the gripper after closure. The free-point error measures how accurately the rest of the object follows its reference trajectory. Fig.~\ref{fig:supp_selected_pointcloud} extends main-paper Fig.~5(a) to additional scenes at matched viewpoints and scales. The cyan boxes mark the contact neighborhoods. The surrounding point clouds show how the predictions extend from each grasp to the rest of the object, so both regions can be examined in the same view.

\subsection{Video Prediction}
With the same TGP and video generator as PointWorld + RAVA, rehearsal reduces 2D trajectory error from 9.58 to 8.12\,px and centroid displacement error from 55.7 to 38.0\,mm (main-paper Tab.~2). The first metric compares point tracks in the image plane. The second compares the displacement of the reconstructed object center with its reference displacement in 3D. Fig.~\ref{fig:supp_case16_case12} extends main-paper Fig.~5(b) to a paper-roll move and a kettle placement. The full frames show the object pose in each scene. The matched-scale crops allow a closer comparison of the object and gripper positions in each prediction against the same reference frame.

\subsection{Material and Viewpoint Editing}
\begin{figure*}[p]
\centering
\input{fig/supp_case16_case12}
\par\vspace{3pt}
\begin{minipage}{\textwidth}
\centering
\captionsetup{type=figure,skip=2pt}
\includegraphics[width=0.82\textwidth]{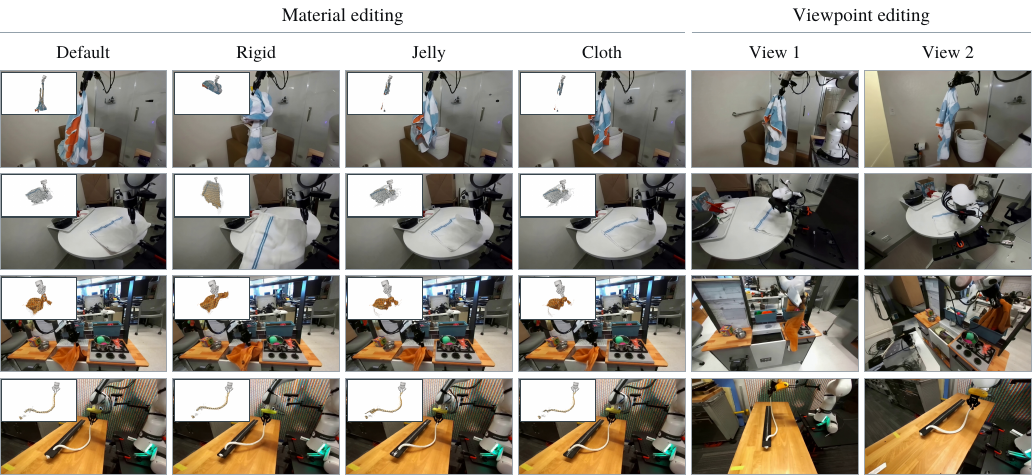}
\caption{\textbf{Material and viewpoint editing across scenes.} Each row shows the same post-grasp instant within a scene. The first four columns fix the camera and show the corresponding meshes. The last two fix the material and change the camera path. The second row shows close-ups of the default, jelly, and cloth meshes.}
\label{fig:supp_video_all}
\end{minipage}
\par\vspace{3pt}
\input{fig/supp_viewpoint_results}
\end{figure*}

Main-paper Fig.~4 tracks three surface points under the same robot action, and Fig.~7 compares the generated appearance. Fig.~\ref{fig:supp_video_all} extends this comparison to additional scenes at a matched post-grasp instant. With the camera fixed, the mesh accompanying each generated frame shows how shape and drape vary with material. With the material fixed, the two views show the same interaction state along different camera paths. Fig.~\ref{fig:supp_trajectories} tracks eight points in two planes at a common scale to separate overall motion from local deformation. Fig.~\ref{fig:supp_viewpoint_results} follows each grasp at three shared frame indices along both paths. A material edit resimulates the object under the original robot action. A viewpoint edit reprojects the existing trajectories along the new camera path. Both edits use the same generator to turn the projected motion into video.

\clearpage

%% file: fig/supp_opening.tex
\par\vspace{4pt}
\begin{minipage}{\textwidth}
\centering
\includegraphics[width=\linewidth]{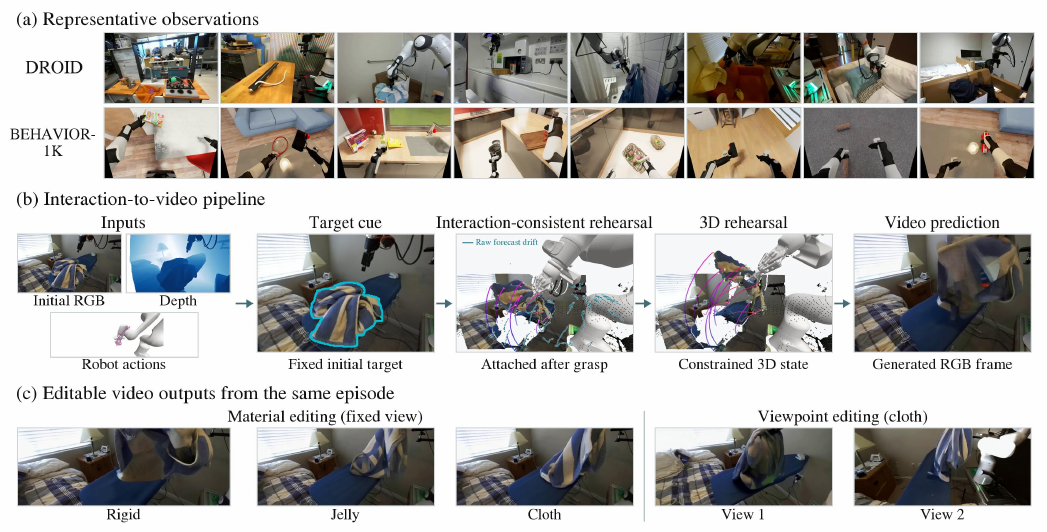}
\captionsetup{skip=4pt}
\captionof{figure}{\textbf{Representative observations and interaction-to-video generation.} (a) Representative initial observations from DROID and BEHAVIOR-1K. (b) RGB-D and robot actions, target cues, interaction-consistent 3D rehearsal, and video generation. (c) Material and viewpoint edits for the same episode.}
\label{fig:supp_opening}
\end{minipage}
\par\vspace{6pt}

%% file: fig/supp_training_settings.tex
\begin{table}[t]
\centering
\caption{\textbf{Training and video settings.}}
\label{tab:supp_training}
\fontsize{8.5}{10.2}\selectfont
\setlength{\tabcolsep}{3pt}
\renewcommand{\arraystretch}{1.10}
\begin{tabular*}{\columnwidth}{@{\extracolsep{\fill}}ll@{}}
\toprule
Parameter & Setting \\
\midrule
\multicolumn{2}{@{}l@{}}{\textit{RAOG}} \\
Training data & 15,000 DROID sequences \\
Objective & Focal loss ($\gamma=2$), type cross-entropy \\
Optimizer / learning rate & AdamW / $3\times10^{-4}$ \\
Epochs / batch size & 1 / 8 \\
\midrule
\multicolumn{2}{@{}l@{}}{\textit{RAVA}} \\
Training data & 3,000 render--RGB pairs \\
& 2,700 DROID, 300 BEHAVIOR-1K \\
Objective & Flow matching \\
Optimizer / learning rate & AdamW / $2\times10^{-5}$ \\
Training / warmup steps & 2,400 / 20 \\
LoRA rank / scale & 64 / 64 \\
Precision / gradient clip & BF16 / 1 \\
Weight decay & 0 \\
\midrule
\multicolumn{2}{@{}l@{}}{Video: 61 frames, $480\times832$ at 15 fps, $T=60$.} \\
\multicolumn{2}{@{}l@{}}{Frozen: PointWorld, video backbone, and VAE.} \\
\bottomrule
\end{tabular*}
\end{table}

%% file: fig/supp_training_supervision.tex
\begin{figure}[t]
\centering
\graphicspath{{assets/}}
\resizebox{\columnwidth}{!}{\input{fig/supp_training_supervision_panel}}
\caption{\textbf{Training examples.} (a) Visible cup outline and responsive object points. (b) Reference 3D trajectories drive the coarse TGP renders in the top row. Colored lines mark projected surface-point tracks. The bottom row shows recorded RGB at the same frames and viewpoints. The initial RGB frame supplies appearance conditioning separately.}
\label{fig:supp_training_supervision}
\end{figure}

%% file: fig/supp_training_supervision_panel.tex
\begingroup
\definecolor{responsive}{RGB}{238,119,31}
\definecolor{backgroundpoints}{RGB}{44,170,182}
\definecolor{targetoutline}{RGB}{225,151,39}
\begin{tikzpicture}[x=1bp,y=-1bp,
  picture/.style={anchor=north west,inner sep=0pt,outer sep=0pt},
  heading/.style={anchor=north west,inner sep=0pt,font=\bfseries\fontsize{9}{11}\selectfont},
  label/.style={anchor=center,inner sep=0pt,font=\fontsize{8}{10}\selectfont},
  rowlabel/.style={label,anchor=west,align=left},
  pairlabel/.style={rowlabel,font=\fontsize{7}{8.5}\selectfont}]
\node[heading] at (0,0) {(a) RAOG point supervision};
\node[label] at (53,17) {Initial DROID RGB};
\node[label] at (183,17) {Target outline and response points};
\node[picture] at (0,28) {\includegraphics[width=106bp]{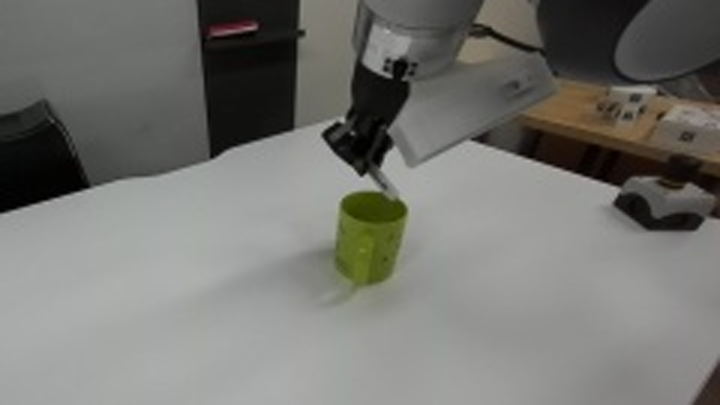}};
\node[picture] at (130,28) {\includegraphics[width=106bp]{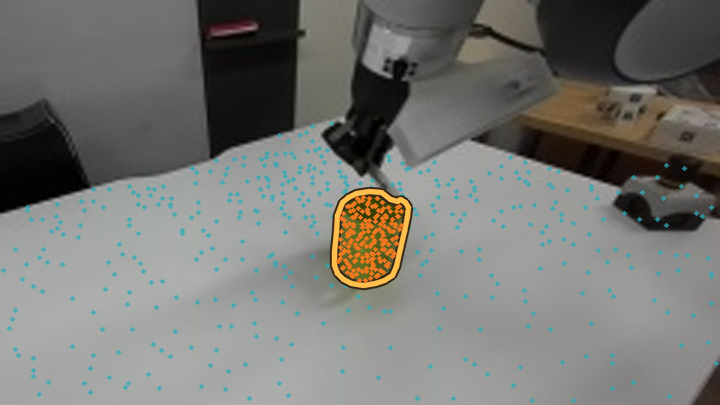}};
\draw[-{Latex[length=4bp,width=3.5bp]},line width=.6bp,black!60] (111,57.8) -- (125,57.8);
\draw[targetoutline,line width=.9bp] (12,100) -- (19,100);
\node[rowlabel] at (23,100) {Target outline};
\fill[responsive] (88,100) circle (1.6bp);
\node[rowlabel] at (94,100) {Responsive points};
\fill[backgroundpoints] (171,100) circle (1.6bp);
\node[rowlabel] at (177,100) {Static points};

\draw[black!21,line width=.45bp] (.25,112) -- (235.75,112);
\node[heading] at (0,119) {(b) RAVA render--RGB pairs};
\foreach \x/\frameindex in {41/10,107/48,173/55}{
  \node[label] at (\x+31.5,138) {\(t=\frameindex\)};
  \node[picture] at (\x,145) {\includegraphics[width=63bp]{rava_render_\frameindex.jpg}};
  \node[picture] at (\x,185) {\includegraphics[width=63bp]{rava_target_\frameindex.jpg}};
}
\node[pairlabel] at (0,163) {TGP render\\Condition};
\node[pairlabel] at (0,203) {Recorded\\RGB target};
\end{tikzpicture}
\endgroup

%% file: fig/supp_structure_detail.tex
\begin{figure}[!t]
\centering
\includegraphics[width=\linewidth]{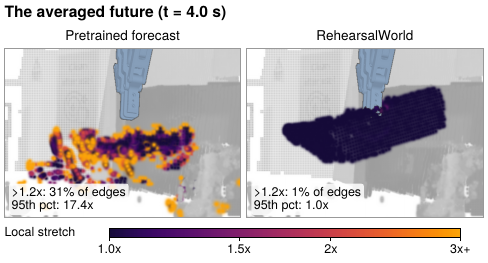}
\caption{\textbf{Local stretch in the forecast and rehearsal.}}
\label{fig:supp_state_detail}
\end{figure}

%% file: fig/supp_material_trajectories.tex
\begin{figure*}[t]
\centering
\includegraphics[width=0.85\textwidth]{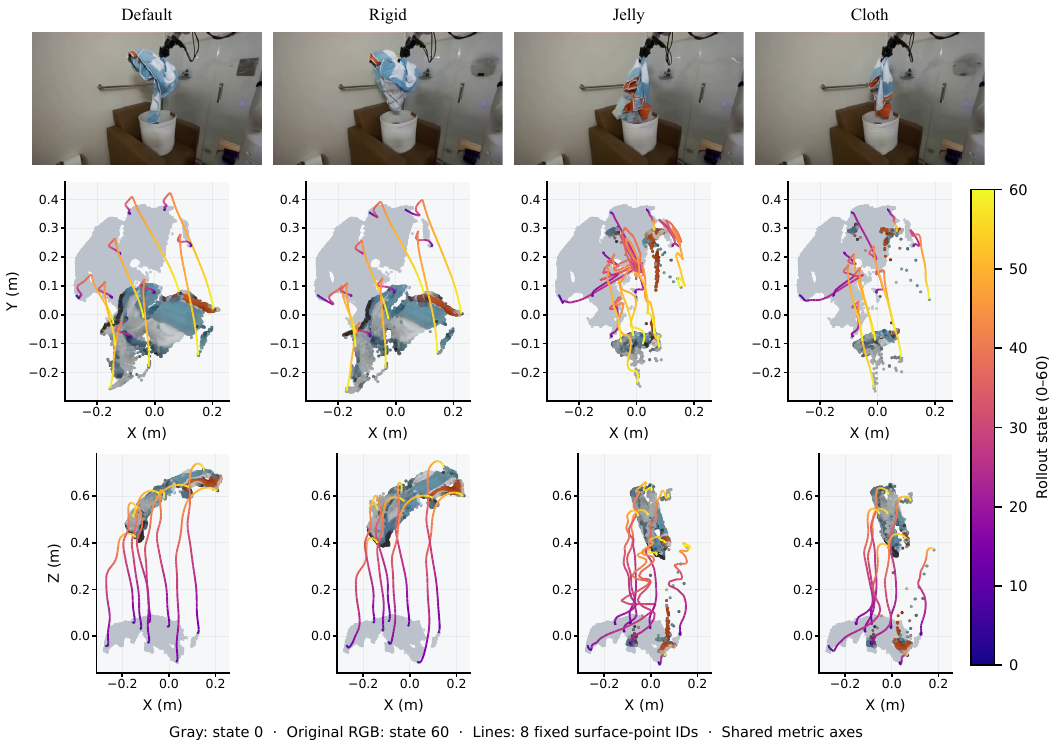}
\caption{\textbf{Generated appearance and surface trajectories across materials.} The four columns show default, rigid, jelly, and cloth. The top row shows generated RGB at frame 58. The lower rows plot eight fixed surface points over states 0--60 on the X--Y and X--Z planes, with common metric axes. Gray and original point colors mark states 0 and 60, and curve color encodes state.}
\label{fig:supp_trajectories}
\end{figure*}

%% file: fig/controlled_physical_ablation.tex
\begin{table*}[!t]
\centering
\caption{\textbf{Ablations of material constraints and trajectory guidance.} We compare constraint variants and guidance weights with the same initial state and actions. Rankings use unrounded values.}
\label{tab:supp_physical_ablation}
\label{tab:supp_physical_controls}
\fontsize{8.5}{10}\selectfont
\setlength{\tabcolsep}{4pt}
\renewcommand{\arraystretch}{1.12}
\begin{tabular*}{\textwidth}{@{\extracolsep{\fill}}lccccc@{}}
\toprule
Setting & \shortstack{Cloth edge\\strain $\downarrow$} & \shortstack{Cloth overstretch\\(\%) $\downarrow$} & \shortstack{Jelly overstretch\\(\%) $\downarrow$} & \shortstack{Support-plane\\penetration (\%) $\downarrow$} & \shortstack{Motion error\\(mm) $\downarrow$} \\
\midrule
Full solver ($w=0.003$) & 0.18 & \second{0.89} & \second{15.09} & 0.07 & \second{12.50} \\
w/o grasp attachment & \best{0.14} & 1.37 & 24.54 & \best{0.00} & 118.60 \\
w/o shape, stretch, and long-range constraints & 2.35 & 28.43 & 26.03 & \second{0.04} & 25.40 \\
w/o support-plane collision and friction & \second{0.18} & 0.93 & 15.25 & 0.67 & 15.20 \\
w/o trajectory guidance ($w=0$) & 0.23 & \best{0.29} & \best{5.86} & 0.07 & 192.40 \\
\midrule
\multicolumn{6}{@{}l}{\emph{Guidance weight sweep}} \\
$w=0.03$ & 0.22 & 3.52 & 18.33 & 0.07 & 48.60 \\
$w=0.3$ & 0.44 & 9.06 & 25.73 & 0.42 & \best{10.20} \\
\bottomrule
\end{tabular*}
\end{table*}

%% file: fig/supp_selected_pointcloud.tex
\begin{minipage}{\textwidth}
\centering
\captionsetup{type=figure,skip=4pt}
\begingroup
\resizebox{0.95\textwidth}{!}{%
\begin{tikzpicture}[x=1bp,y=-1bp]
\node[anchor=north west,inner sep=0] at (0,0)
{\includegraphics[width=504bp]{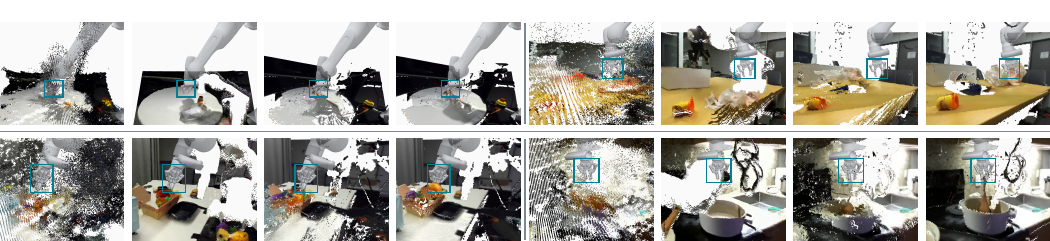}};
\tikzset{
  comparisonmethod/.style={
    anchor=north,inner sep=0,align=center,yshift=-2.5bp,
    font=\rmfamily\mdseries\fontsize{8.8}{10}\selectfont},
  comparisonmethodbold/.style={
    comparisonmethod,font=\rmfamily\bfseries\fontsize{8.8}{10}\selectfont}
}
\node[comparisonmethod] at (29.750000,0) {Kinema4D\,\cite{kinema4d2026}};
\node[comparisonmethod] at (93.250000,0) {TesserAct\,\cite{zhen2025tesseract}};
\node[comparisonmethod] at (156.750000,0) {PointWorld\,\cite{huang2026pointworld}};
\node[comparisonmethodbold] at (220.250000,0) {RehearsalWorld};
\node[comparisonmethod] at (283.750000,0) {Kinema4D\,\cite{kinema4d2026}};
\node[comparisonmethod] at (347.250000,0) {TesserAct\,\cite{zhen2025tesseract}};
\node[comparisonmethod] at (410.750000,0) {PointWorld\,\cite{huang2026pointworld}};
\node[comparisonmethodbold] at (474.250000,0) {RehearsalWorld};
\end{tikzpicture}%
}
\endgroup
\caption{\textbf{Qualitative 3D interaction comparisons across scenes.} Cyan boxes mark contact neighborhoods. We back-project TesserAct point clouds from generated depth.}
\label{fig:supp_selected_pointcloud}
\end{minipage}

%% file: fig/supp_case16_case12.tex
\begin{minipage}{\textwidth}
\centering
\captionsetup{type=figure,skip=2pt}
\begingroup
\resizebox{0.82\textwidth}{!}{%
\begin{tikzpicture}[x=1bp,y=-1bp]
\node[anchor=north west,inner sep=0] at (0,0)
{\includegraphics[width=508bp]{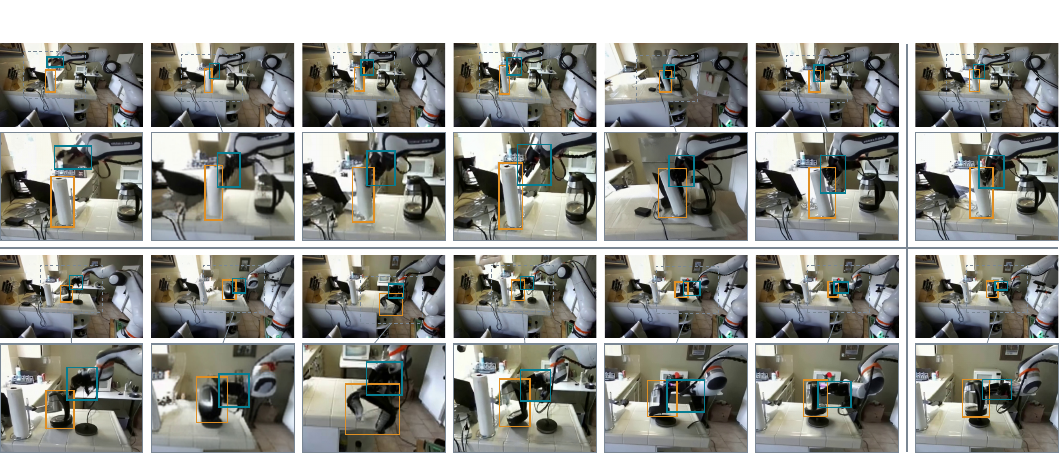}};
\tikzset{
  comparisonheading/.style={
    anchor=west,inner sep=0,yshift=-2.5bp,
    font=\rmfamily\mdseries\fontsize{8.8}{10}\selectfont},
  comparisonmethod/.style={
    anchor=north,inner sep=0,align=center,yshift=-2.5bp,
    font=\rmfamily\mdseries\fontsize{8.8}{10}\selectfont},
  comparisonmethodlong/.style={
    comparisonmethod,font=\rmfamily\mdseries\fontsize{8}{9}\selectfont},
  comparisonmethodbold/.style={
    comparisonmethod,font=\rmfamily\bfseries\fontsize{8.8}{10}\selectfont}
}
\node[comparisonheading] at (0,4.6) {Video prediction};
\node[comparisonmethod] at (34.285714,10) {Wan2.1-I2V\,\cite{wan2025}};
\node[comparisonmethod] at (106.857143,10) {Ctrl-World\,\cite{guo2025ctrlworld}};
\node[comparisonmethod] at (179.428571,10) {Kinema4D\,\cite{kinema4d2026}};
\node[comparisonmethod] at (252.000000,10) {TesserAct\,\cite{zhen2025tesseract}};
\node[comparisonmethodlong] at (324.571429,10) {\mbox{PointWorld\,\cite{huang2026pointworld} + RAVA}};
\node[comparisonmethodbold] at (397.142857,10) {RehearsalWorld};
\node[comparisonmethodbold] at (473.714286,10) {GT};
\end{tikzpicture}%
}
\endgroup
\caption{\textbf{Additional qualitative video prediction comparisons.} Generated and reference frames (GT) for two scenes appear in rows 1 and 3, with matched-scale close-ups in rows 2 and 4. Orange boxes mark the target objects, and cyan boxes mark the grippers.}
\label{fig:supp_case16_case12}
\end{minipage}

%% file: fig/supp_viewpoint_results.tex
\begin{minipage}{\textwidth}
\centering
\captionsetup{type=figure,skip=2pt}
\includegraphics[width=0.82\textwidth]{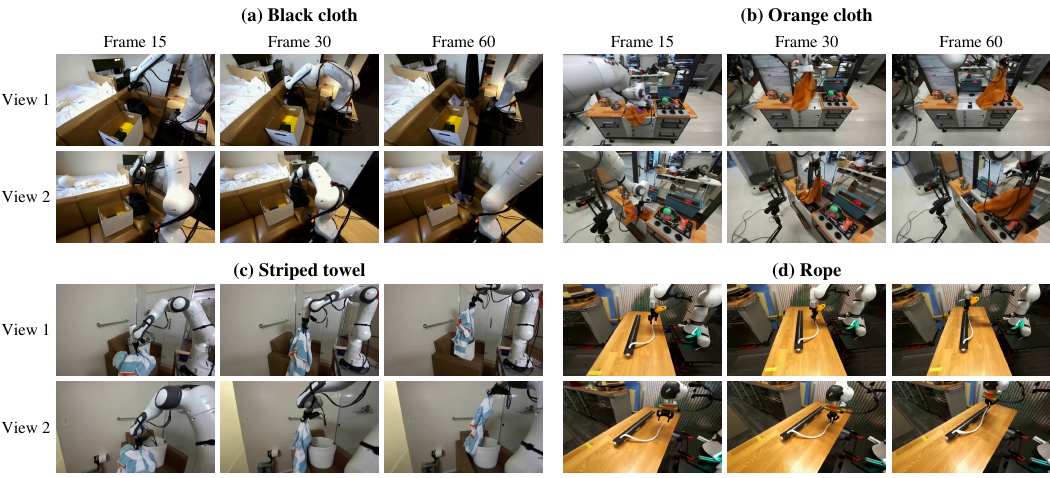}
\caption{\textbf{Generated sequences under viewpoint edits.} The four scenes show (a) black cloth, (b) orange cloth, (c) a striped towel, and (d) a white cord. Within each scene, rows show two camera paths and columns show source frames 15, 30, and 60 (zero-indexed).}
\label{fig:supp_viewpoint_results}
\end{minipage}